\newif\ifCORR
\CORRtrue
\ifCORR
\documentclass{tlp-corr}
\else
\documentclass{tlp}
\fi
\makeatletter
\usepackage{amsfonts}
\usepackage{amssymb}
\usepackage{amstext}
\usepackage{aopmath}
\usepackage{enumerate}
\usepackage{graphicx}
\usepackage{verbatim}
\usepackage{mathrsfs}
\usepackage{afterpage}
\usepackage{multicol}
\usepackage[justification=justified,singlelinecheck=true]{caption}
\usepackage{url}
\usepackage{float}
\usepackage{xspace}
\graphicspath{{figures/}}

\newcommand{\nop}[1]{}
\newcommand{\memph}[1]{\emph{#1}\xspace}

\newcommand{\href}[1]{{#1}\xspace} 
 
\newcommand{\citep}[1]{\cite{#1}}
\newcommand{\citealp}[1]{\citeNP{#1}}
\newcommand{\citeauthorandyear}[1]{\citeN{#1}}
\newcommand{\citeauthor}[1]{\citeANP{#1}}
\newcommand{\be}{\begin{enumerate}}
\newcommand{\ee}{\end{enumerate}}
\newcommand{\bex}[1]{\begin{example}\label{#1}}
\newcommand{\bexcite}[2]{\begin{example}[\hspace{-0.005cm}{\cite{#2}}]\label{#1}}
\newcommand{\eex}{\end{example}}
\newcommand{\beq}[1]{\begin{equation}\label{#1}}
\newcommand{\eeq}{\end{equation}}
\newcommand{\bi}{\begin{itemize}}
\newcommand{\ei}{\end{itemize}}
\newcommand{\bd}[1]{\begin{definition}\label{#1}}
\newcommand{\bdcite}[2]{\begin{definition}[{\citealp{#2}}]\label{#1}}
\newcommand{\bdcitedef}[3]{\begin{definition}[{\citealp{#2}, {#3}}]\label{#1}}
\newcommand{\ed}{\end{definition}}
\newcommand{\bp}{\begin{proof}}
\newcommand{\ep}{\end{proof}}

\newenvironment{sourcebox}[2]{title={\rm\progLabel{#2}} \hfill {\rm #1}}{}
\newcommand{\bsource}[2]{\begin{sourcebox}{#1}{#2}}
\newcommand{\esource}[0]{\end{sourcebox}}
\newcommand{\mathverb}[1]{\mbox{\texttt{#1}}}

 \newcommand{\tool}[1]{\texttt{#1}\xspace}
 \newcommand{\sealion}{\tool{SeaLion}}
 \newcommand{\ouroboros}{\tool{Ouroboros}}
 \newcommand{\clasp}{\tool{Clasp}}

 \newcommand{\dlv}{\tool{DLV}}

 \newcommand{\dlvhex}{\tool{DLVHEX}}

 \newcommand{\gringo}{\tool{Gringo}}

 \newcommand{\kara}{\tool{Kara}}

 \newcommand{\wasp}{\tool{WASP}}
 \newcommand{\spock}{\tool{Spock}}
 
 \newcommand{\aspviz}{\tool{ASPVIZ}}
 \newcommand{\idpdraw}{\tool{IDPDraw}}
 \newcommand{\lonsdaleite}{\tool{Lonsdaleite}}

\newcommand{\Pol}{{\rm P}\xspace}
\newcommand{\NP}{\mbox{\rm NP}\xspace}
\newcommand{\SigmaP}[1]{\ensuremath{{\Sigma}_{#1}^{\Pol}}\xspace}

\newcommand{\namesym}[1]{\textsc{#1}}

\newcommand{\csym}{\namesym{C}}
\newcommand{\cprogram}{\csym-program\xspace}
\newcommand{\cprogramo}{\csym-program}
\newcommand{\crule}{\csym-rule\xspace}
\newcommand{\cruleo}{\csym-rule}
\newcommand{\cprograms}{\csym-programs\xspace}
\newcommand{\crules}{\csym-rules\xspace}
\newcommand{\catom}{\csym-atom\xspace}
\newcommand{\catomo}{\csym-atom}
\newcommand{\catoms}{\csym-atoms\xspace}
\newcommand{\cfact}{\csym-fact\xspace}
\newcommand{\cfacts}{\csym-facts\xspace}
\newcommand{\cliteral}{\csym-literal\xspace}
\newcommand{\cliteralo}{\csym-literal}
\newcommand{\cliterals}{\csym-literals\xspace}

\newcommand{\domain}[1]{\ensuremath{D_{#1}}\xspace}
\newcommand{\satisfiers}[1]{\ensuremath{C_{#1}}\xspace}
\newcommand{\compl}[1]{\ensuremath{\bar{#1}}}
\newcommand{\pnf}[1]{\ensuremath{{#1}^{+}}}
\newcommand{\posOcc}[1]{\ensuremath{\mathit{posOcc}({#1})}\xspace}
\newcommand{\depG}[1]{\ensuremath{G({#1})\xspace}}
\newcommand{\rdepG}[1]{\ensuremath{G_{\mathfrak{R}}({#1})\xspace}}
\newcommand{\depGrel}[0]{\ensuremath{\prec_{\mathfrak{D}}}\xspace}
\newcommand{\topOrder}[0]{\ensuremath{\prec}\xspace}
\newcommand{\rdepGrel}[0]{\ensuremath{\prec_{\mathfrak{R}}}\xspace}

\newcommand{\reductFLP}[2]{\ensuremath{{#1}^{#2}}\xspace}

\newcommand{\setOfAtoms}{\ensuremath{\mathscr{A}}\xspace}

\newcommand{\unfounded}[0]{\ensuremath{\Upsilon}\xspace}
\newcommand{\negpart}[1]{\ensuremath{{#1}^{-}}\xspace}
\newcommand{\stateprog}[1]{\ensuremath{{\progC}_{#1}}\xspace}
\newcommand{\stateI}[1]{\ensuremath{{\inter}_{#1}}\xspace}
\newcommand{\statenegI}[1]{\ensuremath{{\negpart{\inter}}_{#1}}\xspace}
\newcommand{\stateunfounded}[1]{\ensuremath{{\unfounded}_{#1}}\xspace}
\newcommand{\newRule}[2]{\ensuremath{\mathit{\rl_{\mathit{new}}(}{#1,#2}\mathit{)}}\xspace}
\newcommand{\computation}{\ensuremath{\mathrm{C}}\xspace}

\newcommand{\ruleG}[0]{\ensuremath{\rho}\xspace}

 \newcommand{\naf}{\ensuremath{\mathit{not}}\xspace}
 \newcommand{\la}{\ensuremath{\leftarrow}\xspace}
 \newcommand{\body}[1]{\ensuremath{\mathrm{B(}{#1}\mathrm{)}}\xspace}
 \newcommand{\head}[1]{\ensuremath{\mathrm{H(}{#1}\mathrm{)}}\xspace}
 \newcommand{\pbody}[1]{\ensuremath{\mathrm{{B^{+}}(}{#1}\mathrm{)}}\xspace}
 \newcommand{\nbody}[1]{\ensuremath{\mathrm{{B^{-}}(}{#1}\mathrm{)}}\xspace}
 \newcommand{\prog}{\ensuremath{P}\xspace}
 
 \newcommand{\progAG}{\prog}
 
 \newcommand{\progC}{\prog}
 \newcommand{\progCex}[1]{\ensuremath{\progC_{Ex\ref{#1}}}\xspace}
 \newcommand{\rl}{\ensuremath{r}\xspace}
 \newcommand{\AS}[1]{\ensuremath{\mathit{AS}(#1)}\xspace}

\newcommand{\inter}{\ensuremath{I}\xspace}
\newcommand{\proj}[2]{\ensuremath{{#1}|_{#2}}\xspace}
\newcommand{\tuple}[1]{\ensuremath{\langle {#1} \rangle}\xspace}

\newcommand{\abs}[1]{\ensuremath{|{#1}|}\xspace}
\newcommand{\modelsDisj}[0]{\ensuremath{\models^{\exists}}\xspace}

\newcommand{\cond}[0]{\ensuremath{\star}\xspace}

\newcommand{\iec}[0]{i.e.,\xspace}
\newcommand{\cf}[0]{cf.\xspace}

\newcommand{\egc}[0]{e.g.,\xspace}

\newcounter{programnr}
\newcommand{\progLabel}[1]{\refstepcounter{programnr}\label{#1}\progRef{#1}}
\newcommand{\progRef}[1]{\ensuremath{\Pi_{\ref{#1}}}\xspace}

\newtheorem{definition}{Definition}

\newtheorem{example}{Example}

\newtheorem{innercustomthm}{Theorem}
\newenvironment{theoremnr}[1]
  {\renewcommand\theinnercustomthm{#1}\innercustomthm}
  {\endinnercustomthm}

\title{Stepwise Debugging of Answer-Set Programs}
\author[J.~Oetsch, J.~P\"uhrer, and H.~Tompits]{JOHANNES OETSCH, J\"ORG P\"UHRER, AND HANS TOMPITS\thanks{This work was partially supported by the Austrian Science Fund (FWF)
under project P21698.} \\ 
Technische Universit\"at Wien,\\
Institut f\"ur Informationssysteme~184/3,\\
Favoritenstrasse 9-11, A-1040 Vienna,
Austria,\\
\email{\{oetsch,puehrer,tompits\}@kr.tuwien.ac.at}}

\author[J.~Oetsch, J.~P\"uhrer, and H.~Tompits]
{JOHANNES OETSCH\\\\
Technische Universit\"at Wien,\\
Institut f\"ur Informationssysteme~184/3,\\
Favoritenstrasse 9-11, A-1040 Vienna,
Austria,\\
\email{johannes.oetsch@tuwien.ac.at}
\and J\"ORG P\"UHRER\\\\
Universit\"at Leipzig,\\
Institut f\"ur Informatik,\\
Augustusplatz 10, D-04109 Leipzig,
Germany,\\
\email{puehrer@informatik.uni-leipzig.de}
\and
HANS TOMPITS\\\\
Technische Universit\"at Wien,\\
Institut f\"ur Informationssysteme~184/3,\\
Favoritenstrasse 9-11, A-1040 Vienna,
Austria,\\
\email{tompits@kr.tuwien.ac.at}}

\begin{document}

\maketitle
                                              
\begin{abstract}

We introduce a \memph{stepping methodology} for answer-set programming (ASP) that allows for debugging answer-set programs and is based on the stepwise application of rules.
Similar to debugging in imperative languages,
where the behaviour of a program is observed during a step-by-step execution,
stepping for ASP allows for observing the effects that rule applications have in the computation of an answer set.
While the approach is inspired from debugging in imperative programming, it is conceptually different to stepping in other paradigms due to
non-determinism and declarativity that are inherent to ASP.
In particular, unlike statements in an imperative program that are executed following a strict control flow, there is no predetermined order in which to consider rules in ASP during a computation.
In our approach, the user is free to decide which rule to consider active in the next step following his or her intuition.
This way, one can focus on interesting parts of the debugging search space. Bugs are detected during stepping by revealing differences between the actual semantics of the program and the expectations of the user.
As a solid formal basis for stepping, we develop a framework of computations for answer-set programs. For fully supporting different solver languages, we build our framework on an abstract ASP language that is sufficiently general to capture different solver languages. To this end, we make use of abstract constraints as an established abstraction for popular language constructs such as aggregates.
Stepping has been implemented in \sealion, an integrated development environment for ASP. 
We illustrate stepping using an example scenario
and discuss the stepping plugin of \sealion.
Moreover, we elaborate on methodological aspects and the embedding of stepping in the ASP development process. 

\ifCORR
{\it Under consideration in Theory and Practice of Logic Programming (TPLP).}
\fi
\end{abstract}

\section{Introduction}
Answer-set programming (ASP)~\cite{N99,MT99} is a paradigm for declarative problem solving that is popular
amongst researchers in artificial intelligence and knowledge representation.
Yet it is rarely used by software engineers outside academia so far.
Arguably, one obstacle preventing developers from using ASP is a lack of support
tools for developing answer-set programs.
One particular problem in the context of programming support is \memph{debugging of answer-set programs}.
Due to the fully declarative semantics of ASP, it can be quite tedious to detect an error in an answer-set program.
In recent years, debugging in ASP has received some attention~\cite{BD05,S06,BGPSTW07,P07,GPST08,GPSTW09,PSE09,OPT10,OPT10b,OPT11,OPT12,PFSF13,FPF13,S15}.
These previous works are important contributions towards ASP development support,
however current approaches come with limitations to their practical applicability.
First, existing techniques and tools  \memph{only capture a basic ASP language fragment} that does not include many
language constructs that are available and frequently used in modern ASP solver languages, \egc,
aggregates or choice rules are not covered by current debugging strategies 
(with the exception of the work by \citeauthorandyear{PFSF13}, where cardinality constraints are dealt with by translation).
Second,  \memph{usability aspects are often not considered} in current approaches, in particular, the programmer is required to either
provide a lot of data to a debugging system or he or she is confronted with a huge amount of information from the
system (tackling this problem in query-based debugging has been addressed by~\citeauthorandyear{S15}).

This paper introduces a \memph{stepping methodology for ASP},
which is a novel technique for debugging answer-set programs
that is general enough to \memph{deal with current ASP solver languages} and is
\memph{intuitive and easy to use}.
Our method is similar in spirit to 
the widespread and effective debugging strategy in imperative programming, where
the idea is to gain insight into the behaviour of a program by executing statement by statement
following the program's control flow.
In our approach, we allow for stepwise constructing interpretations
by considering
rules of an answer-set program at hand  in a successive manner.
This method guarantees that either an answer set will be reached, or
some error will occur that provides hints why the semantics of the program differs from the user's expectations.
A major difference to the imperative setting is that, due to its declarativity,
ASP lacks any control flow.
Instead, we allow the user to follow his or her intuition on
which rule instances to become active.
This way, one can focus on interesting parts of the debugging search space from the beginning.
For illustration, 
the following answer-set program has $\{a\}$ as its only answer set.
\begin{verbatim}
a :- not b.
b :- not a.
a :- b.
\end{verbatim}
Let's step through the program to obtain explanations why this is the case.
In the beginning of a stepping session, no atom is considered to be true. 
Under this premise, the first two rules are active.
The user decides which of these rules to apply. 
Choosing a rule to be applied in this manner is considered a \memph{step} in our approach.
In case the user chooses the first rule, the atom \verb|a| is derived. 
Then, no further rule is active and one of the answer sets, \verb|{a}| has been reached.
If, on the other hand, the user chooses the second rule in the first step, atom \verb|b| is derived and \verb|a| is considered false.
Then, the third rule becomes active.
However, this rule would derive \verb|a| that is already considered false when choosing the second rule.
In this case, the user sees that no answer set can be reached based on the initial choice.

Besides single steps that allow the user to consider one rule instance at a time, we also lay the ground for so-called \memph{jumps}.
The intuition is that in a jump multiple rule instances and even multiple non-ground rules can be considered at once.
Jumping significantly speeds up the stepping process which makes our technique a usable tool for debugging in practice.
Consider the following encoding of an instance of the three-colouring problem in the \gringo language~\cite{GKKS11}:\begin{verbatim}
1{color(X,red;green;blue)}1 :- node(X).

:- edge(X,Y), color(X,C), color(X,C).

node(X):-edge(X,Y).
node(Y):-edge(X,Y).
edge(1,2). edge(1,3). edge(1,4).
edge(2,4). edge(2,5). edge(2,6).
edge(3,4). edge(3,5). edge(3,6).
edge(4,5). edge(5,6).
\end{verbatim}
The user expects the program to have answer sets but it does not.
Following our approach, the reason for that can be found after two actions.
First, trusting the ``instance'' part of the program, the user applies a jump on all rules of this part,
and, intuitively, gets all atoms implied by these rules as an intermediate result.
Second, the user applies an arbitrary instance of the rule 
\begin{verbatim} 
1{color(X,red;green;blue)}1 :- node(X).
\end{verbatim}
that is active under the atoms derived during the jump.
Suppose, the user chooses the instance
\begin{verbatim} 
1 {color(1, red), color(1, green), color(1, blue)} 1 :- node(1).
\end{verbatim} 
and selects \verb|color(1, red)| to be true.
Then, the debugging system reveals that the instance
\begin{verbatim} 
:- edge(1, 2), color(1, red), color(1, red).
\end{verbatim} 
of the ``check'' constraint becomes unexpectedly active.
Now, the users sees that the second occurrence of \verb|color(X,C)| in the constraint has to be replaced by \verb|color(Y,C)|.
Generally, bugs can be detected whenever stepping reveals
differences between the actual semantics of the program and the expectations of the user.

In order to establish a solid formal basis for our stepping technique,
we developed a framework of computations for answer-set programs.
For fully supporting current solver languages, we were faced with several challenges.
For one, the languages of answer-set solvers differ from each other and from formal ASP languages
in various ways.
In order to develop a method that works for different solvers, we need an abstract ASP language that
is sufficiently general to capture actual solver languages.
To this end, we make use of abstract constraints~\citep{MR04,MT04} as an established
abstraction for language constructs such as aggregates, weight constraints, and external atoms.
We rely on a semantics for arbitrary abstract-constraint programs with disjunctions
that we introduced for this purpose in previous work~\cite{OPT12b}.
In contrast to other semantics for this type of programs, it is
compatible with the semantics of all the ASP solvers we want to support, namely, \clasp~\cite{gekasc12c}, \dlv~\cite{LPFEGPS06}, and \dlvhex~\cite{Redl16}.
Note that our framework for computations for abstract-constraint programs differs from the one by \citeauthorandyear{LPST10}.
We did not build on this existing notion for three reasons. First, it does not cover rules with disjunctive heads which we want to support.
Second, steps in this framework correspond to the application of multiple rules. Since our method is rooted in the analogy to stepping in procedural languages, where an ASP rule corresponds to a statement in an imperative language, we focus on steps corresponding to application of a single rule. 
Finally, the semantics of non-convex literals differs from that of \dlvhex in the existing approach. 
A thorough discussion on the relation of the two notions of computations is given in Section~\ref{sec:related}.

Another basic problem deals with the grounding step in which variables are removed from
answer-set programs before solving.
In formal ASP languages, the grounding of a program consists of all rules resulting from substitutions of variables by ground terms. 
In contrast, actual grounding tools apply many different types of simplifications and pre-evaluations for creating a variable-free program.
In order to close this gap between formal and practical ASP, \citeauthor{Puehrer14} developed abstractions of the grounding step together with an abstract notion of non-ground answer-set program as the base language for the stepping methodology in his PhD thesis~\citep{Puehrer14}.
Based on that, stepping can easily be applied to existing solver languages and it becomes robust to changes to these languages.
As we focus on the methodological aspects of stepping in this article, we do not present these abstractions and implicitly use grounding as carried out by actual grounding tools.

The stepping technique has been implemented in \sealion~\cite{OPT13}, an integrated development environment for ASP.
We discuss how \sealion can be used for stepping answer-set programs written in the \gringo or the \dlv language.\footnote{The framework introduced in this paper subsumes and significantly extends previous versions of the stepping technique for normal logic programs \citep{OPT10b,OPT11} and DL-programs \citep{OPT12}.}

\subsection*{Outline}
Next, we provide the formal background that is necessary for our approach.
We recall the syntax of disjunctive abstract-constraint programs and the semantics on which we base our framework~\cite{OPT12b}.
Section~\ref{sec:computation} introduces a framework of computations that allows for breaking the semantics down to the level of individual rules.
After defining states and computations, we show several properties of the framework, most importantly
soundness and completeness in the sense that 
the result of a successful computation is an answer set and that
every answer set can be constructed with a computation.
Moreover, we study language fragments for which a simpler form of computation suffices.
In Section~\ref{sec:stepping}, we
present the stepping technique for debugging answer-set programs based on our computation framework.
We explain \memph{steps} and \memph{jumps}
as a means to progress in a computation using an example scenario.
Moreover, we discuss methodological aspects of stepping on the application level (how stepping is used for 
debugging and program analysis) and the top level (how stepping is embedded in the ASP development process).
We illustrate the approach with several use cases and describe the stepping interface of \sealion.
Related work is discussed in Section~\ref{sec:related}. We compare stepping to other debugging approaches for ASP and discuss the relation of our computation framework to that of~\citeauthorandyear{LPST10} and transition systems for ASP~\cite{L11,LierlerT16,BrocheninLM14}.
We conclude the paper in Section~\ref{sec:conclusion}.

In \ifCORR \ref{sec:guidelines} \else supplementary Appendix A \fi 
we compile guidelines for stepping and give general recommendations for ASP development.
Selected and short proofs are included in the main text and all remaining proofs are provided in \ifCORR \ref{sec:proofs}\else supplementary Appendix~B\fi. \section{Background}\label{sec:background}

As motivated in the introduction, we represent grounded answer-set programs by abstract-constraint programs~\citep{MR04,MT04,OPT12b}.
Non-ground programs will be denoted by programs in the input language of \gringo.
Thus, we implicitly assume that grounding translates (non-ground) \gringo rules to rules of abstract-constraint programs. 
For a detailed formal account of our framework in the non-ground setting we refer the interested reader to the dissertation of \citeauthorandyear{Puehrer14}.

We assume a fixed set \setOfAtoms of ground atoms.
\bd{def:interpretation}
An \memph{interpretation} is a set $\inter\subseteq\setOfAtoms$ of ground atoms.
A ground atom $a$ is \memph{true} under interpretation \inter,
symbolically $\inter\models a$, if
$a\in \inter$, otherwise it is \memph{false} under \inter.
\ed
We will use the symbol $\not\models$ to denote the complement of a relation denoted with the symbol $\models$ in different contexts.

For better readability, we sometimes make use of the following notation when the reader may interpret the intersection of two sets $\inter$ and $X$ of ground atoms as a projection from \inter to $X$.
\bd{def:projection}
For two sets $\inter$ and $X$ of ground atoms, $\proj{\inter }{X}=\inter \cap X$ is the \memph{projection} of $\inter$ to~$X$.
\ed

\subsection{Syntax of Abstract-Constraint Programs}\label{sec:progC}
Rule heads and bodies of abstract-constraint programs
are formed by so-called \memph{abstract-constraint atoms}.

\bdcite{def:catom}{MR04,MT04}
An \memph{abstract con\-straint}, \memph{abstract-\-constraint atom}, or \memph{\catomo}, is a pair $A=\tuple{D,C}$,
where $D\subseteq\setOfAtoms$ is a finite set called the \memph{domain} of $A$, denoted by $\domain{A}$,
and $C\subseteq 2^D$ is a collection of sets of ground atoms, called the \memph{satisfiers} of $A$, denoted by $\satisfiers{A}$.
\ed

We can express atoms also as \catoms.
In particular,
for a ground atom $a$, we identify the \catom $\tuple{\{a\},\{\{a\}\}}$ with $a$.
We call such \catoms  \memph{elementary}.

We will also make use of default negation
in abstract-constraint programs.
An \memph{ab\-stract-constraint literal}, or \memph{\cliteralo}, is a \catom $A$ or a
default negated \catom $\naf~A$.

Unlike the original definition, we introduce 
abstract-constraint programs with disjunctive rule heads.

\bd{def:crule}
An \memph{abstract-constraint rule}, or simply \memph{\cruleo}, is an expression of the form
\beq{crule}
A_{1} \lor \cdots \lor A_{k} \la A_{k+1}, \ldots, A_{m}, \naf\ A_{m+1}, \ldots, \naf\ A_{n}\ , 
\eeq
where $0 \leq k \leq m \leq n$ and any $A_{i}$, for $1\leq i\leq n$, is a \catom.
\ed
Note that if all disjuncts share the same domain they can be expressed by a single \catom (see~\citealp{Puehrer14})
but in general disjunction adds expressivity.

We identify different parts of a \crule
and introduce some syntactic properties.
\bd{def:rulePartsC}
For a \crule $r$ of form~(\ref{crule}),
\bi
\item $\body{r} = \{A_{k+1}, \ldots, A_{m},\naf\ A_{m+1}, \ldots, \naf\ A_{n}\}$ is the \memph{body} of $r$,
\item $\pbody{r} = \{A_{k+1}, \ldots, A_{m}\}$ is the \memph{positive body} of $r$,
\item $\nbody{r} = \{A_{m+1}, \ldots, A_{n}\}$ is the \memph{negative body} of $r$, and 
\item $\head{r} = \{A_{1}, \ldots, A_{k}\}$ is the \memph{head} of $r$.
\ei
\ed
\noindent
If $\body{r}=\emptyset$ and $\head{r}\neq\emptyset$, then $r$ is a \memph{\cfact}.
For \cfacts, we
usually omit the symbol ``$\la$''.
A \crule $r$  of form~(\ref{crule}) is \memph{normal} if $k=1$ and \memph{positive} if $m=n$.

We define the domain of a default negated \catom $\naf~A$ as
$\domain{\naf\ \! A}=\domain{A}$.
Then, the domain $\domain{S}$ of a set $S$ of \cliterals is given by
\[\domain{S}=\bigcup_{L \in S} \domain{L}.\]
\noindent
Finally, the domain of a \crule $r$ is 
\[\domain{r}=\bigcup_{X\in\head{r}\cup\body{r}}\domain{X}.\]

\bd{def:programsAndClassesC}
An \memph{abstract-constraint program}, or simply \memph{\cprogramo}, is a finite set of \crules.
A \cprogram is \memph{normal}, respectively \memph{positive}, if it contains only normal, respectively positive, \crules.
A \cprogram is \memph{elementary} if it contains only elementary \catoms.
\ed

\subsection{Satisfaction Relation}\label{sec:satC}
Intuitively, a \catom $\tuple{D,C}$ is a literal whose truth depends on the truth of all atoms in $D$ under a given interpretation.
The satisfiers in $C$ explicitly list which combinations of true atoms in $D$ make the \catom true.
\bd{def:satCatom}
An interpretation $\inter$ \emph{satisfies} a \catom $\tuple{D,C}$, symbolically $I \models \tuple{D,C}$, if 
$\proj{\inter}{D}\in C$.
Moreover, $I \models \naf\  \tuple{D,C}$ iff 
$I \not\models \tuple{D,C}$.
\ed

Important criteria for distinguishing classes of \catoms
are concerned with their semantic behaviour with respect to growing (or shrinking) 
interpretations.
In this respect, we identify monotonicity properties
in the following.

\bd{def:monotonicity}
A \cliteral $L$ is \emph{monotone} if, for all interpretations $\inter$ and $\inter'$, if $\inter\subseteq \inter'$ and
$\inter\models L$, then also $\inter'\models L$.
$L$ is \emph{convex} if, for all interpretations $\inter$, $\inter'$, and $\inter''$, if $\inter\subseteq \inter'\subseteq \inter''$, $\inter\models L$, and $\inter''\models L$, then also $\inter'\models L$.
Moreover, a \cprogram $\progC$ is monotone (respectively, convex) if for all $r\in\progC$ all \cliterals $L\in\head{r}\cup\body{r}$ are monotone (respectively, convex).
\ed
\noindent
Next, the notion of satisfaction is extended to \crules and \cprograms
in the obvious way.
\bd{def:satCprograms}
An interpretation $\inter$ satisfies a set $S$ of \cliterals, symbolically $\inter\models S$, if $\inter\models L$ for all $L\in S$.
For brevity, we will use the notation $\inter\modelsDisj S$ to denote that
$\inter\models L$ for some $L\in S$.
Moreover, $\inter$ satisfies a \crule $r$, symbolically $\inter\models r$, if 
$\inter\models\body{r}$ implies $\inter\modelsDisj\head{r}$.
A \crule $r$ such that $\inter\models\body{r}$ is called \emph{active under $\inter$}.
As well, $\inter$ satisfies a set $\prog$ of \crules, symbolically $\inter\models \prog$, if
$\inter\models r$ for every $r\in\prog$.
If $\inter\models \prog$, we say that $\inter$ is a \emph{model} of $\prog$.
\ed

\subsection{Viewing ASP Constructs as Abstract Constraints}\label{sec:agexweightAsC}
We want to use abstract con\-straints as a uniform means to represent 
common constructs in ASP solver languages.
As an example, we recall how  weight constraints~\cite{SNS02}
can be expressed as \catoms. 
In a similar fashion, we can use them as abstractions of \egc aggregates~\cite{FLP04,FPL11}
or external atoms~\cite{EIST05}.
Note that the relation between abstract constraints and ASP constructs is well known
and motivated abstract constraints in the first place (cf.~\citealp{MR04,MT04}).

\bdcite{def:weightConstraint}{SNS02}
A \memph{weight constraint} is an expression of form 
\beq{eq:weight-con}
l\ [a_1=w_1,\dots,a_k=w_k,\naf\ a_{k+1}=w_{k+1},\dots,\naf\ a_n=w_n]\ u\,,
\eeq
where each $a_i$ is a ground atom and each weight $w_i$ is a real number, for $1\leq i \leq n$.
The lower bound $l$ and the upper bound $u$ are either a real number, $\infty$, or $-\infty$.
\ed
For a weight constraint to be true,
the sum of weights $w_i$ of those atoms $a_i$, $1 \leq i \leq k$, that are true and the weights of the atoms $a_i$, $k < i \leq n$, that are false must lie within the lower and the upper bound.
Thus, a weight constraint of form~(\ref{eq:weight-con})
corresponds to the \catom $\tuple{D,C}$,
where 
the domain 
$D=\{a_1,\dots,a_n\}$
consists of the atoms appearing in the weight constraint
and 
\[
C=\{X\subseteq D\mid l\leq (\sum_{1\leq i \leq k,a_i\in X} w_i +  \sum_{k < i \leq n,a_i\not\in X} w_i) \leq u\}\,.
\]
\noindent

\subsection{Semantics and Characterisations based on External Support and Unfounded Sets}
The semantics we use~\cite{OPT12b} extends the FLP-semantics~(\citeauthor{FLP04}~2004;~2011)
 and coincides with the original notion of answer sets by \citeauthorandyear{GL91} on many important classes
of logic programs, including elementary \cprograms.
Similar to the original definition of answer sets,
\citeauthor{FPL11} make use of a program reduct depending on a candidate interpretation \inter
for determining whether \inter satisfies a stability criterion and thus is considered an answer set.
However, the reduct of Faber, Leone, and Pfeifer differs in spirit from that of Gelfond and Lifschitz
as it does not reduce the program to another syntactic class (the 
Gelfond-Lifschitz reduct of an elementary \cprogram is always positive).
Instead, the so-called \memph{FLP-reduct}, defined next, keeps the individual rules intact
and just ignores all rules that are not active under the candidate interpretation.
\bd{def:reductFLPc}
Let \inter be an interpretation, and let \progC be a \cprogram.
The FLP-reduct of \progC with respect to \inter is given by 
$
\reductFLP{\progC}{\inter}=\{\rl \in \progAG \mid \rl\mbox{ is active under }\inter\}
$.
\ed

The notion of answer sets for abstract-constraint programs defined next
provides the semantic foundation for the computation model we use for debugging.
\begin{definition}[\citealp{OPT12b}]\label{def:answersets}
Let $\progC$ be a \cprogram, and let $\inter$ be an interpretation.
$\inter$ is an \emph{answer set} of $\progC$ if 
$\inter\models\progC$ and there is no $\inter'\subset\inter$ such that ${\progC}$, ${\inter}$, and ${\inter'}$ satisfy the following condition:
\bi\item[(\cond)] 
for every $r\in\reductFLP{\progC}{\inter}$ with $\inter'\models\body{r}$, there is some $A\in\head{r}$ with $\inter'\models A$ and 
$\proj{\inter'}{\domain{A}}=\proj{\inter}{\domain{A}}$.
\ei
The set of all answer sets of $\progC$ is denoted by $\AS{\progC}$.
\end{definition}
\noindent
The purpose of Condition (\cond) is to prevent minimisation within \catoms: the requirement $\proj{\inter'}{\domain{A}}=\proj{\inter}{\domain{A}}$ ensures
that a satisfier $\{a,b\}$ can enforce $b$ to be true in an answer set even if the same \catom has a satisfier $\{a\}$.
As a consequence answer sets need not be subset minimal (see~\citealp{Puehrer14} for details).

Our choice of semantics has high solver compatibility as its objective 
as we want to support \gringo, \dlv, and \dlvhex.
We need an FLP-style treatment of non-convex literals for being compatible with \dlvhex,
disjunctions to support \dlv and \dlvhex, and
we must allow for weight constraints in rule heads for compatibility with \gringo.
Note that \gringo/\clasp treats aggregates in the way suggested by~\citeauthorandyear{F11}.
As a consequence, its semantics differs from our semantics in some cases, when recursion is used through negated c-atoms,  
as ours is an extension of the FLP semantics. For an in-depth comparison of FLP-semantics and Ferraris semantics we refer to work by \citeauthorandyear{T10}.
An example where the semantics differ is given by the single rule \gringo program
\verb|a :- not 0{a}0|
that has only the empty set as answer set under our semantics, whereas \clasp also admits
$\{a\}$ as an answer set.
In practice, this difference only hardly influences the compatibility with \gringo, as aggregates are seldom used in this way. 
We examined all \gringo encodings send to the second ASP competition and could not find any such usage.

Our framework of computations for stepping 
is based on a characterisation of the semantics of Definition~\ref{def:answersets}
in terms of \memph{external supports}.
Often, answer sets are computed following a two-step strategy: First, a model of the program is built, and second, it is checked whether
this model obeys a foundedness condition ensuring that it is an answer set.
Intuitively, every set of atoms in an answer set must be ``supported'' by
some active rule that derives one of the atoms.
Here, it is important that the reason for this rule to be active does not
depend on the atom it derives.
Such rules are referred to as \emph{external support}~\citep{L05b}.
The extension of this notion to our setting is the following.
\begin{definition}[\citealp{OPT12b}]\label{def:extsupp}
Let $r$ be a \crule, $X$ a set of atoms, and $\inter$ an interpretation. Then, $r$ is an \emph{external support for $X$ with respect to $\inter$} if
\begin{itemize}
\item[(i)] $\inter\models\body{r}$,
\item[(ii)] $\inter\setminus X \models\body{r}$,
\item[(iii)] there is some $A\in\head{r}$ with
$\proj{X}{\domain{A}}\neq\emptyset$ and $\proj{\inter}{\domain{A}}\subseteq S$, for some $S\in\satisfiers{A}$, and
\item[(iv)] for all $A\in\head{r}$ with $\inter\models A$, $\proj{(X\cap\inter)}{\domain{A}}\neq\emptyset$ holds.
\end{itemize}
\end{definition}
\noindent
Condition~(i) ensures that $r$ is active. Condition~(ii) prevents self-support by guaranteeing the support to be ``external'' of $X$, \iec $r$ is also be active without the atoms in $X$.
In case \inter is a model, Items (iii) and (iv) jointly ensure that there is some
\catom $A$ in the head of $r$ that is satisfied by $\inter$
and derives some atom of $X$.

We can express the absence of an external support in an interpretation by the concept of an \emph{unfounded set}.
\begin{definition}[\citealp{OPT12b}]
Let $\progC$ be a \cprogram, $X$ a set of atoms, and $\inter$ an interpretation.
Then, $X$ is \emph{unfounded in $\progC$ with respect to $\inter$}
if there is 
no \crule $r\in\progC$ that is an external support for $X$ with respect to $\inter$.
\end{definition}

\begin{corollary}[\citealp{OPT12b}]\label{cor:asunfounded}
Let $\progC$ be a \cprogram and $\inter$ an interpretation.
Then, $\inter$ is an answer set of $\progC$ iff
$\inter$ is a model of $\progC$ and
there is no set $X$ with $\emptyset\subset X\subseteq \inter$
that is unfounded in $\progC$ with respect to $\inter$.
\end{corollary}

 \section{Computation Framework}\label{sec:computation}

In this section, we want to break the conceptual complexity
of the semantics down to artefacts the programmer is familiar with: the rules the user has written or, more precisely, their ground instances.
To this end, we introduce a framework of computations that
captures the semantics described in the previous section.
In this computation model, on top of which we will introduce stepping in Section~\ref{sec:stepping}, an interpretation is built up step-by-step
by considering an increasing number of rule instances to be active.
A \memph{computation} in our framework is a sequence of \memph{states}
which are structures that keep information which rules and atoms have already been
considered and what truth values were assigned to those atoms.
Utilising the framework, 
only one rule and the atoms it contains have to be considered at once
while building up an interpretation until an answer set is reached
or a source for the unexpected behaviour becomes apparent.

In the next two subsections, we introduce states and computations.
In Section~\ref{sec:computationProperties}, we define and show
some properties of computations that we need later on when we describe stepping.
Section~\ref{sec:stableComputations} is concerned with the existence of a stable computation which is a simpler form of computation that suffices for
many popular classes of answer-set programs.
We discuss existing work related to our computation framework later in Section~\ref{sec:related}.

\subsection{States}\label{sec:state}
Our framework is based on sequences
of states, reassembling computations, in which an increasing number of
ground rules are considered
that build up a monotonically growing interpretation.
Besides that interpretation, states also capture literals which cannot become true in subsequent steps
and sets that currently lack external support in the state's interpretation.
\begin{definition}\label{def:state}
A \memph{state structure} $S$ is a tuple $\tuple{\progC,I,\negpart{I},\unfounded}$,
where $\progC$ is a set of  \crules, $I$ is an interpretation, \negpart{I} a set of atoms such that $I$ and \negpart{I} are disjoint, and \unfounded is a collection of sets of atoms.
We call $\domain{S}=I\cup\negpart{I}$ the \memph{domain} of $S$
and define $\stateprog{S}=\progC$, $\stateI{S}=I$, $\statenegI{S}=\negpart{I}$, and
$\stateunfounded{S}=\unfounded$.

A state structure $\tuple{\progC,I,\negpart{I},\unfounded}$ is a \memph{state} if
\bi
\item[(i)] $I\models\body{r}$ and $I\modelsDisj\head{r}$ for every $r\in\progC$, 
\item[(ii)] $\domain{r}\subseteq \domain{S}$ for every $r\in\progC$,
and 
\item[(iii)] $\unfounded=\{X\subseteq I\mid \mbox{$X$ is 
unfounded in $\progC$ with respect to $I$}\}$.
\ei
We call $\tuple{\emptyset,\emptyset,\emptyset,\{\emptyset\}}$ the \memph{empty state}.
\end{definition}
\noindent
Intuitively, we use the first component \progC of a state to collect \crules that
the user has considered to be active and satisfied.
The interpretation $I$ collects atoms that have been considered true.
Condition~(i) ensures that \progC and $I$ are compatible in the sense that 
every \crule that is considered active and satisfied is active and satisfied with respect to $I$.
Dual to $I$, the interpretation \negpart{I} collects atoms that the user has considered to be false.
We require that all atoms appearing in a \crule in \progC is either in $I$ or in \negpart{I}
which is expressed in Condition~(ii).
Finally, the set $\unfounded$ keeps track of unfounded subsets of $I$, as stated in Condition~(iii).
Intuitively, as we will see later, when building a computation, 
the aim is to get rid of all unfounded sets (except for the empty set) in order to compute an answer set
of a \cprogram.
If a state does not contain such unfounded sets then we call it stable:
\bd{def:stable}
A state $S$ is \memph{stable} if $\stateI{S}\in\AS{\stateprog{S}}$.
\ed
\noindent 
The intuition is that when a state $S$ is stable, no more \crules need to be added
to \stateprog{S} to provide missing external support for the atoms in the current interpretation \stateI{S}.
Note that a state $S$ is stable exactly when $\stateunfounded{S}=\{\emptyset\}$.
For example, the empty state is a stable state.

\bex{ex:state}

Consider the \crules
\[
\begin{array}{l@{~}r@{\ \leftarrow\ }l}
r_1:& \tuple{\{a,b\},\{\emptyset,\{a\},\{b\},\{a,b\}} & \naf\ a\\
r_2:&b & a
\end{array}
\]
\noindent
and the state structures
\[
\begin{array}{r@{\ = }l@{\qquad}r@{\ = }l}
S_1 & \tuple{\{r_1\},\emptyset,\{a,b\},\{\emptyset\}},&
S_2 & \tuple{\{r_1\},\{b\},\{a\},\{\emptyset\}},\\
S_3 & \tuple{\{r_1\},\{a,b\},\emptyset,\{\emptyset\}},&
S_4 & \tuple{\{r_2\},\{a,b\},\emptyset,\{\emptyset\}},\\
S_5 & \tuple{\{r_2\},\{a,b\},\emptyset,\{\{b\},\{a,b\}\}}.
\end{array}
\]
\noindent
$S_1$ and $S_2$ are stable states.
$S_3$ is not a state as $\stateI{S_3}\not\models \body{r_1}$.
$S_4$ is not a state as the sets $\{b\}$ and $\{a,b\}$ are unfounded in $\stateprog{S_4}$ with respect to $\stateI{S_4}$ but $\{b\}\not\in\stateunfounded{S_4}$ and $\{a,b\}\not\in\stateunfounded{S_4}$.
$S_5$ is a state but not stable.
\eex

\subsection{Computations}
Next, we show how we can proceed forward in a computation, i.e.,
which states might follow a given state using a successor relation for state structures.
\begin{definition}~\label{def:successor}
For a state $S=\tuple{\progC,I,\negpart{I},\unfounded}$ and a state structure $S'=\tuple{\progC',I',\negpart{I'},\unfounded'}$,
$S'$ is a \memph{successor of $S$} if there is a \crule
$r\in\progC'\setminus\progC$ and sets $\Delta,\negpart{\Delta}\subseteq\domain{r}$ such that
\bi
\item[(i)]$\progC'=\progC\cup\{r\}$,
\item[(ii)]$I'=I\cup\Delta$, $\negpart{I'}=\negpart{I}\cup\negpart{\Delta}$, and $\domain{S}\cap(\Delta\cup\negpart{\Delta})=\emptyset$,
\item[(iii)] $\domain{r}\subseteq\domain{S'}$,
\item[(iv)]$I \models \body{r}$, 
\item[(v)]$I'\models \body{r}$ and $I'\modelsDisj\head{r}$, and
\item[(vi)] $X'\in\unfounded'$ iff $X'=X \cup \Delta'$, where
                            $X\in\unfounded$,
                            $\Delta'\subseteq\Delta$, and
                            $r$ is not an external support for $X'$ with respect to $I'$.
\ei
We denote $r$ by $\newRule{S}{S'}$.
\end{definition}
\noindent
Condition~(i) ensures that a successor state considers exactly one rule more to be active. Conditions~(ii) and~(iii) express that the interpretations $I$ and $\negpart{I}$
are extended by the so far unconsidered literals in $\Delta$ and $\negpart{\Delta}$ appearing in the new \crule $\newRule{S}{S'}$.
Note that from $S'$ being a state structure we get that 
$\Delta$ and $\negpart{\Delta}$ are distinct.
A requirement for considering $\newRule{S}{S'}$ as next \crule is that it is active under the current interpretation $I$, expressed by Condition~(iv).
Moreover, $\newRule{S}{S'}$ must be satisfied and still be active under the succeeding interpretation, as required by Condition~(v).
The final condition ensures that the unfounded sets of the successor
are extensions of the previously unfounded sets that are not externally supported by 
the new rule.

Here, it is interesting that only extended previous unfounded sets can be unfounded sets in the extended \cprogram $\progC'$
and that $\newRule{S}{S'}$ is the only \crule which could provide external support for them in $\progC'$ with respect to the new interpretation $I'$ as seen next.

\begin{theorem}\label{th:extSuppLocally}
Let $S$ be a state 
and $S'$ a successor of $S$,
where $\Delta=\stateI{S'}\setminus\stateI{S}$. 
Moreover, let $X'$ be a set of literals with $\emptyset\subset X'\subseteq \stateI{S'}$.
Then, the following statements are equivalent:
\bi
\item[{\rm (}i{\rm )}] 
$X'$ is unfounded in $\stateprog{S'}$ with respect to $\stateI{S'}$.
\item[{\rm (}ii{\rm )}] $X'=\Delta'\cup X$, where $\Delta'\subseteq \Delta$, $X\in\stateunfounded{S}$, and
$\newRule{S}{S'}$ is not an external support for $X'$ with respect to $\stateI{S'}$.
\ei
\end{theorem}
\noindent
The result shows that
determining the unfounded sets in
a computation after adding a further \crule $r$
can be done locally, \iec
only supersets of previously unfounded sets
can be unfounded sets, and if such a superset
has some external support
then it is externally supported by $r$.
The result also implies that the successor relation suffices to ``step'' from one state to another.
\begin{corollary}\label{cor:successorstate}
Let $S$ be a state and $S'$ a successor of $S$.
Then, $S'$ is a state.
\end{corollary}
\bp
We show that the Conditions (i), (ii), and (iii) of Definition~\ref{def:state} hold for $S'$.
Consider some rule $r\in\stateprog{S'}$. 
In case $r=\newRule{S}{S'}$,
$\stateI{S'}\models\body{r}$ and $\stateI{S'}\modelsDisj\head{r}$ hold because of Item~(v) of Definition~\ref{def:successor} and
$\domain{r}\subseteq \domain{S'}$ because of Item~(iii) of the same definition.
Moreover, in case $r\neq\newRule{S}{S'}$, we have $r\in\stateprog{S}$.
As $S$ is a state, we have $\domain{r}\subseteq\domain{S}$.
Hence, since $\domain{S}\subseteq\domain{S'}$ also $\domain{r}\subseteq\domain{S'}$.
Note that $\proj{\stateI{S'}}{\domain{r}}=\proj{\stateI{S}}{\domain{r}}$ because of 
Item~(ii) of Definition~\ref{def:successor}.
Therefore, as $\stateI{S}\models\body{r}$ and $\stateI{S}\modelsDisj\head{r}$, also
$\stateI{S'}\models\body{r}$ and $\stateI{S'}\modelsDisj\head{r}$.
From these two cases, we see that Conditions (i) and (ii) of Definition~\ref{def:state} hold for $S'$.
Finally, Condition~(iii) follows from Item~(vi) of Definition~\ref{def:successor} and Theorem~\ref{th:extSuppLocally}.
\ep

Next, we define computations based on the notion of a state.
\begin{definition}\label{def:computation}
A \memph{computation} is a sequence $C=S_0,\dots,S_n$ of states such that $S_{i+1}$ is a successor of $S_i$, for all $0\leq i < n$.
We call $C$ \memph{rooted} if $S_0$ is the empty state
and \memph{stable} if each $S_i$ is stable, for $0\leq i \leq n$.
\end{definition}

\subsection{Properties}\label{sec:computationProperties}
We next define when a computation has failed, gets stuck, is complete, or has succeeded.
Intuitively, failure means that the computation reached a point where no answer set
of the \cprogram can be reached.
A computation is stuck when the last state activated rules
deriving literals that are inconsistent with previously chosen active rules.
It is considered complete when there are no more unconsidered active rules.
Finally, a computation has succeeded if an answer set has been reached.
\begin{definition}\label{def:computationsemantics2}
Let $\progC$ be a \cprogram and $C=S_0,\dots,S_n$ a computation such that $\stateprog{S_n}\subseteq\progC$.
Then, $C$ is called a \memph{computation for} $\progC$. Moreover,
\begin{itemize}
\item $C$ has \memph{failed for $P$ at step $i$} if there is no answer set $I$ of $\progC$ such that
$\stateI{S_{i}}\subseteq I$, $\statenegI{S_{i}}\cap I=\emptyset$,
and $\stateprog{S_{i}}\subseteq\reductFLP{\progC}{\inter}$;
\item is \memph{complete for $P$} if for every rule $r\in\reductFLP{\progC}{\stateI{S_{n}}}$, we have $r\in\stateprog{S_{n}}$;
\item is \memph{stuck in $P$}
if it is not complete for $P$ but there is no successor $S_{n+1}$ of $S_n$ such that $\newRule{S_n}{S_{n+1}}\in\progC$;
\item \memph{succeeded for $P$} if it is complete 
and $S_n$ is stable.
\end{itemize}
\end{definition}
\bex{ex:computation}
Let $\progCex{ex:computation}$ be the \cprogram consisting of the \crules
\[
\begin{array}{l@{~}r@{\ \leftarrow\ }l}
r_1:& a & \tuple{\{a,b\},\{\emptyset,\{a,b\}\}}\\
r_2:&b & a\\            r_3:&a & b\\
r_4:&\tuple{\{c\},\{\emptyset,\{c\}\}}&\\
r_5:&&c
\end{array}
\]
\noindent 
that has $\{a,b\}$ as its single answer set,
and consider the sequences
\[
\begin{array}{rl@{}l}
\bullet&C_1=&\tuple{\emptyset,\emptyset,\emptyset,\{\emptyset\}},\\&&
    \tuple{\{r_4\},\{\},\{c\},\{\emptyset\}},\\&&
    \tuple{\{r_4,r_1\},\{a,b\},\{c\},\{\{a\},\{b\}\}},\\[4pt]
\bullet&C_2=&\tuple{\emptyset,\emptyset,\emptyset,\{\emptyset\}},
    \tuple{\{r_4\},\{\},\{c\},\{\emptyset\}},
    \tuple{\{r_4,r_1\},\{a,b\},\{c\},\{\{a\},\{b\}\}},\\&&
    \tuple{\{r_4,r_1,r_2\},\{a,b\},\{c\},\{\{a\}\}},
    \tuple{\{r_4,r_1,r_2,r_3\},\{a,b\},\{c\},\{\emptyset\}},\\[4pt]
\bullet&C_3=&\tuple{\{r_4,r_1,r_2,r_3\},\{a,b\},\{c\},\{\emptyset\}},\\[4pt]
\bullet&C_4=&\tuple{\emptyset,\emptyset,\emptyset,\{\emptyset\}},
    \tuple{\{r_4\},\{c\},\emptyset,\{\emptyset\}},\\[4pt]
\bullet&C_5=&\tuple{\{r_4,r_1,r_2,r_3\},\{a,b,c\},\emptyset,\{\emptyset\}},\\[4pt]
\bullet&C_6=&\tuple{\{r_5\},\emptyset,\{c\},\{\emptyset\}},\quad \mbox{and}\\[4pt]
\bullet&C_7=&\tuple{\emptyset,\emptyset,\emptyset,\{\emptyset\}}, 
    \tuple{\{r_4,r_1\},\{a,b\},\{c\},\{\{a\},\{b\}\}}.\\[4pt]
\end{array}
\]
\noindent
$C_1$, $C_2$, $C_3$, $C_4$, and $C_5$ are computations for $\progCex{ex:computation}$.
The sequence $C_6$ is not a computation, as $\tuple{\{r_5\},\emptyset,\{c\},\{\emptyset\}}$ is not a state.
$C_7$ is not a computation, as the second state in $C_7$ is not a successor of the empty state.
$C_1$, $C_2$, and $C_4$ are rooted.
$C_3$, $C_4$, and $C_5$ are stable.
$C_2$ and $C_3$ are complete and have succeeded for $\progCex{ex:computation}$.
$C_1$ is complete for $\progCex{ex:computation}\setminus\{r_2,r_3\}$
but has failed for $\progCex{ex:computation}\setminus\{r_2,r_3\}$ at Step $0$
because $\progCex{ex:computation}\setminus\{r_2,r_3\}$ has no answer set.
$C_4$ has failed for $\progCex{ex:computation}$ at Step $1$.
$C_5$ has failed for $\progCex{ex:computation}$ at Step $0$ and is stuck in $\progCex{ex:computation}$.
\eex

The following result guarantees the soundness of our framework of computations.
\begin{theorem}\label{th:sound}
Let $\progC$ be a \cprogram
and $C\!=\!S_0,\dots,S_n$ a computation that has succeeded for $\progC$.
Then, $\stateI{S_n}$ is an answer set of $\progC$.
\end{theorem}
\bp
As $C$ is complete for $\progC$, we have $\reductFLP{\progC}{\stateI{S_n}}\subseteq
\stateprog{S_n}$.
Conversely, we have 
$\stateprog{S_n}\subseteq\reductFLP{\progC}{\stateI{S_n}}$
because for each $r\in\stateprog{S_n}$
we have $r\in\progC$ and $\stateI{S_n}\models\body{r}$.
By stability of $S_n$, we get that $\stateI{S_n}\in\AS{\stateprog{S_n}}$.
The conjecture holds since then $\stateI{S_n}\in\AS{\reductFLP{\progC}{\stateI{S_n}}}$.
\ep
\noindent
The computation model is  also complete in the following sense:
\begin{theorem}\label{th:complete}
Let $S_0$ be a state, $\progC$ a \cprogram with $\stateprog{S_0}\subseteq\progC$,
and $I$ an answer set of $\progC$ with $\stateI{S_0}\subseteq I$ and $I\cap\statenegI{S_0}=\emptyset$.
Then, there is a computation $S_0,\dots,S_n$ that has succeeded for $\progC$ such that
$\stateprog{S_n}=\reductFLP{\progC}{I}$ and $\stateI{S_n}=I$.
\end{theorem}
\noindent
As the empty state, $\tuple{\emptyset,\emptyset,\emptyset,\{\emptyset\}}$, is trivially a  state, we can make the completeness aspect of the previous result more apparent in the following corollary:
\begin{corollary}\label{cor:completeness}
Let $\progC$ be a \cprogram and $I\in\AS{\progC}$.
Then, there is a rooted computation $S_0,\dots,S_n$ that has succeeded for $\progC$ such that
$\stateprog{S_n}=\reductFLP{\progC}{\inter}$ and $\stateI{S_n}=I$.
\end{corollary}
\bp
The claim follows immediately from Theorem~\ref{th:complete} in case $S_0=\tuple{\emptyset,\emptyset,\emptyset,\{\emptyset\}}$.
\ep

Note, that there are states that do not result from rooted computations,
\egc the state 
$
\tuple{\{a\la b\},\{a,b\},\emptyset,\{\emptyset,\{a,b\},\{b\}\}}
$
is not a successor of any other state.
However, for stable states, we can guarantee the existence of rooted computations.
\begin{corollary}\label{cor:stablestategrounded}
Let $S$ be a stable state.
Then, there is a rooted computation $S_0,\dots,S_n$ with
$S_n=S$.
\end{corollary}
\bp
The result is a direct consequence of Corollary~\ref{cor:completeness} and Definition~\ref{def:stable}.
\ep

The next theorem lays the ground for the jumping technique that we introduce in Section~\ref{sec:stepping}.
It allows for extending a computation by considering multiple rules of a program at once
and using ASP solving itself for creating this extension.
\begin{theorem}\label{th:jumping}
Let $\progC$ be a \cprogram,
$C=S_0,\dots,S_n$ a computation for $\progC$,
$\progC'$ a set of \crules with $\progC'\subseteq\progC$,
and $I$ an answer set of $\stateprog{S_n}\cup\progC'$
with $\stateI{S_n}\subseteq I$ and $I\cap\statenegI{S_n}=\emptyset$.
Then, there is a computation $C'=S_0,\dots,S_n,S_{n+1},\dots,S_m$ for $\progC$, such that 
$S_m$ is stable, $\stateprog{S_m}=\stateprog{S_n}\cup\reductFLP{\progC'}{\inter}$
and $\stateI{S_m}=I$.
\end{theorem}
\bp
By Theorem~\ref{th:complete},
as
$\stateprog{S_n}\subseteq\stateprog{S_n}\cup\progC'$,
$\stateI{S_n}\subseteq I$, and $I\cap\statenegI{S_n}=\emptyset$,
there is a computation $S_n,\dots,S_m$ that has succeeded for $\stateprog{S_n}\cup\progC'$ such that
$\stateprog{S_m}=\reductFLP{(\stateprog{S_n}\cup\progC')}{I}$ and $\stateI{S_m}=I$.
Then, $S_m$ is stable and,
as $\reductFLP{\stateprog{S_n}}{I}=\stateprog{S_n}$, we have
$\stateprog{S_m}=\stateprog{S_n}\cup\reductFLP{\progC'}{\inter}$.
As $\stateprog{S_m}\subseteq\progC$, we have that
$C'=S_0,\dots,S_n,S_{n+1},\dots,S_m$ is a computation 
for $\progC$.
\ep

The following result illustrates that the direction
one chooses for building up a certain interpretation, \iec the order of the rules considered in a computation, is irrelevant in the sense that eventually the same state will be reached.
\begin{proposition}
Let $\progC$ be a \cprogram and $C=S_0,\dots,S_n$ and $C'=S'_0,\dots,S'_m$ computations complete for $\progC$ such that $S_0=S'_0$.
Then, $I_{S_n}=I_{S'_m}$ iff $S_n=S'_m$ and $n=m$.
\end{proposition}
\bp
The ``if'' direction is trivial.
Let $I=I_{S_n}=I_{S'_m}$.
Towards a contradiction, assume $\stateprog{S_n}\neq\stateprog{S'_m}$.
Without loss of generality, we focus on the case that there is some
$r\in\stateprog{S_n}$ such that $r\not\in\stateprog{S'_m}$.
Then, it holds that $I\models r$, $I\models \body{r}$, and $r\in\progC$.
Consequently, $r\in\reductFLP{\progC}{\inter}$. By completeness of $C'$, we have
$r\in\stateprog{S'_m}$ which contradicts our assumption.
Hence, we have $\stateprog{S_n}=\stateprog{S'_m}$.

By definition of a state, from $\stateI{S_n}=\stateI{S'_m}$ and $\stateprog{S_n}=\stateprog{S'_m}$, it follows that $\stateunfounded{S_n}=\stateunfounded{S'_m}$.
Towards a contradiction, assume $\statenegI{S_n}\neq\statenegI{S'_m}$.
Without loss of generality we focus on the case that there is some
$a\in\statenegI{S_n}$ such that $a\not\in\statenegI{S'_m}$.
Consider the integer $i$ where $0<i\leq n$ such that $a\in\statenegI{S_i}$ but
$a\not\in\statenegI{S_{i-1}}$.
Then, by definition of a successor, for $r=\newRule{S_{i-1}}{S_{i}}$, we have 
$a\in\negpart{\Delta}$ for some $\negpart{\Delta}\subseteq\domain{r}$.
As then $a\in\domain{r}$ and, as $\stateprog{S_n}=\stateprog{S'_m}$,
we have $r\in\stateprog{S'_m}$, it must hold that $a\in\domain{S'_m}$
by definition of a state structure. From $I\cap\statenegI{S_n}=\emptyset$ we know that $a\not\in I$. Therefore, since $a\in I\cup\statenegI{S'_m}$, we get that
$a\in \statenegI{S'_m}$, being a contradiction to our assumption.
As then $S_n=S'_m$, $\stateprog{{S_0}}=\stateprog{{S'_0}}$, and since in every step in a computation
exactly one rule is added it must hold that $n=m$.
\ep

For rooted computations, the domain of each state is determined
by the atoms in the \crules it contains.
\begin{proposition}
Let $C=S_0,\dots,S_n$ be a rooted computation.
Then, $\stateI{S_i}=\proj{\stateI{S_n}}{\domain{\stateprog{S_i}}}$ and
$\statenegI{S_i}=\proj{\statenegI{S_n}}{\domain{\stateprog{S_i}}}$,
for all $0\leq i\leq n$.
\end{proposition}
\bp
The proof is by contradiction.
Let $j$ be the smallest index with
$0\leq j\leq n$ such that
$\stateI{S_j}\neq\proj{\stateI{S_n}}{\domain{\stateprog{S_j}}}$ or
$\statenegI{S_j}\neq\proj{\statenegI{S_n}}{\domain{\stateprog{S_j}}}$.
Note that $0<j$ as $\stateI{S_0}=\statenegI{S_0}=\domain{\stateprog{S_0}}=\emptyset$.
As $S_{j}$ is a successor of $S_{j-1}$, we have 
$\stateI{S_{j}}=\stateI{S_{j-1}}\cup \Delta$ and
$\statenegI{S_{j}}=\statenegI{S_{j-1}}\cup \negpart{\Delta}$,
where
$\Delta,\negpart{\Delta}\subseteq\domain{\newRule{S_{j-1}}{S_{j}}}$,
$\domain{S_{j-1}}\cap(\Delta\cup\negpart{\Delta})=\emptyset$,
and $\domain{\newRule{S_{j-1}}{S_{j}}}\subseteq\stateI{S_{j}}\cup\statenegI{S_{j}}$.
As we have
$\stateI{S_{j-1}}=\proj{\stateI{S_n}}{\domain{\stateprog{S_{j-1}}}}$ and
$\statenegI{S_{j-1}}=\proj{\statenegI{S_n}}{\domain{\stateprog{S_{j-1}}}}$,
it holds that
\[
\begin{array}{c}
\stateI{S_{j-1}}
\cup
\proj{\stateI{S_n}}{\domain{\delta}} 
=\proj{\stateI{S_n}}{\domain{\stateprog{S_{j-1}}}} 
\cup 
\proj{\stateI{S_n}}{\domain{\delta}}
=\proj{\stateI{S_n}}{\domain{\stateprog{S_{j}}}} \mbox{ and}\\
\statenegI{S_{j-1}}
\cup
\proj{\statenegI{S_n}}{\domain{\delta}} 
=\proj{\statenegI{S_n}}{\domain{\stateprog{S_{j-1}}}} 
\cup 
\proj{\statenegI{S_n}}{\domain{\delta}} 
=\proj{\statenegI{S_n}}{\domain{\stateprog{S_{j}}}},
\end{array}
\]
\noindent
where
$\domain{\delta}=\domain{\stateprog{S_{j}}}\setminus\domain{\stateprog{S_{j-1}}}$.
For establishing the contradiction, it suffices to show that
$\proj{\stateI{S_n}}{\domain{\delta}}=\Delta$
and
$\proj{\statenegI{S_n}}{\domain{\delta}}=\negpart{\Delta}$.
Consider some $a\in\Delta$.
Then, $a\in\domain{\delta}$ because $a\in\domain{\newRule{S_{j-1}}{S_{j}}}$, $\domain{S_{j-1}}\cap(\Delta\cup\negpart{\Delta})=\emptyset$,
and $\domain{\stateprog{S_{j-1}}}\subseteq\domain{S_{j-1}}$.
Moreover, $a\in\stateI{S_{j}}$ implies $a\in\stateI{S_{n}}$
and therefore $\Delta\subseteq\proj{\stateI{S_n}}{\domain{\delta}}$.
Now, consider some $b\in\proj{\stateI{S_n}}{\domain{\delta}}$.
As $\domain{\newRule{S_{j-1}}{S_{j}}}\subseteq\stateI{S_{j}}\cup\statenegI{S_{j}}$,
we have $b\in\stateI{S_{j}}\cup\statenegI{S_{j}}$.
Consider the case that $b\in\statenegI{S_{j}}$.
Then, also $b\in\statenegI{S_{n}}$ which is a contradiction to
$b\in\stateI{S_n}$ as $S_n$ is a state structure.
Hence, $b\in\stateI{S_{j}}=\stateI{S_{j-1}}\cup \Delta$.
First, assume 
$b\in\stateI{S_{j-1}}$.
This leads to a contradiction as then $b\in\domain{\stateprog{S_{j-1}}}$
since $\stateI{S_{j-1}}=\proj{\stateI{S_n}}{\domain{\stateprog{S_{j-1}}}}$.
It follows that $b\in\Delta$ and therefore
$\Delta=\proj{\stateI{S_n}}{\domain{\delta}}$.
One can show that
$\negpart{\Delta}=\proj{\statenegI{S_n}}{\domain{\delta}}$ analogously.
\ep

\subsection{Stable Computations}\label{sec:stableComputations}
In this section, we are concerned with the existence of stable computations, \iec
computations that do not involve unfounded sets.
We single out an important class of \cprograms for which one can solely rely on this type of computation and also give examples of \cprograms that do not allow
for succeeding stable computations.

Intuitively, the $\SigmaP{2}$-hardness of the semantics (cf.~\citealp{Puehrer14}), demands for unstable computations in the general case.
This becomes obvious when considering that for a given \cprogram
one could guess a candidate sequence $C$ for a stable computation in polynomial time.
Then, a polynomial number of checks whether each state is a successor of the previous one in the sequence suffices to establish whether $C$ is a computation.
Following Definition~\ref{def:successor}, these checks can be done in polynomial time
when we are allowed to omit Condition~(vi) for unfounded sets.
Hence, answer-set existence for the class of \cprograms for which every answer set can be built up with stable computations is in \NP.

Naturally, it is interesting whether there are syntactic classes of \cprograms
for which we can rely on stable computations only.
It turns out that many syntactically simple \cprograms already require the use of unfounded sets.
\bex{ex:unstableHCNM}
Consider \cprogram~\progCex{ex:unstableHCNM} consisting of the \crules
\[
\begin{array}{l@{~}r@{\ \leftarrow\ }l}
r_1:& a & b \quad \quad \mbox{and}\\
r_2:& b & \tuple{\{a\},\{\emptyset,\{a\}\}}.
\end{array}
\]
\noindent
We have that $\{a,b\}$ is the only answer set of $\progCex{ex:unstableHCNM}$
and 
\[
\begin{array}{r@{}l}
C=&\tuple{\emptyset,\emptyset,\emptyset,\{\emptyset\}},\\&
    \tuple{\{r_2\},\{a,b\},\emptyset,\{\emptyset,\{a\}\}},\\&
    \tuple{\{r_2,r_1\},\{a,b\},\emptyset,\{\emptyset\}}
\end{array}
\]
\noindent
is the only computation that succeeds for $\progCex{ex:unstableHCNM}$: starting at the empty state, only rule $r_2$ is active, thus it must be the new rule in the successor.
When, deciding the truth values for the atoms in $\domain{r_2}$, $r_2$ requires $b$ to be positive, and $a$ must be true as well, as otherwise the computation is stuck due to violation of $r_1$.
The second state of $C$ contains the singleton $\{a\}$ as unfounded set.
\eex
\noindent
As Example~\ref{ex:unstableHCNM} shows, unstable computations are already
required for a \cprogram without disjunction and a single monotone \catom.
Hence, also the use of weaker restrictions, like
convexity of \catoms or
some notion of head-cycle freeness~\citep{BD94}, is not sufficient.

One can observe, that the \cprogram from the example has cyclic positive dependencies between atoms $a$ and $b$.
Hence, we next explore whether such dependencies influence the 
need for computations that are not stable.
To this end, we introduce notions of \memph{positive dependency} in a \cprogram.

\bd{def:depGraph}
Let $S$ be a set of \cliterals. Then, the \memph{positive normal form} of $S$ is given by
$$\pnf{S}=\{A \mid A\in S, A\mbox{ is a \catom}\}\cup \{\compl{A} \mid \naf\ A\in S\},$$ where $\compl{A}=\tuple{\domain{A},2^{\domain{A}}\setminus \satisfiers{\domain{A}}}$ is the \memph{complement} of $A$.
Furthermore, the \memph{set of positive atom occurrences} in $S$
is given by
$
\posOcc{S}=\bigcup_{A\in\pnf{S},X\in\satisfiers{A}}X
$.

Let \progC be a \cprogram.
The \memph{positive dependency graph} of \progC is the directed graph
\[
\depG{\progC}=\tuple{\domain{\progC},\{\tuple{a,b}\mid r\in\progC, a\in\posOcc{\head{r}}, b\in\posOcc{\body{r}}\}}.\]
\ed

We next introduce the notion of \memph{absolute tightness} for describing
\cprograms without cyclic positive dependencies after
recalling basic notions of graph theory.
For a (directed) graph $G=\tuple{V,\prec}$, the \memph{reachability relation} of $G$ is the transitive closure of $\prec$.
Let $\prec'$ be the reachability relation of $G$. 
Then, $G$ is \memph{acyclic} if there is no $v\in V$ such that $v\prec'v$.

\bd{def:abstight}
Let \progC be a \cprogram.
\progC is \memph{absolutely tight} if \depG{\progC} is acyclic.
\ed

One could assume that absolute tightness
paired with convexity or monotonicity is sufficient
to guarantee stable computations because
absolute tightness forbids positive dependencies among disjuncts
and the absence of such dependencies lowers the complexity of elementary \cprograms \citep{BD94}.
However, as the following example illustrates,
this is not the case for general \cprograms.

\bex{ex:unstableTDMa}
Consider \cprogram~\progCex{ex:unstableTDMa} consisting of the \crules
\[
\begin{array}{l@{~}r@{\ \leftarrow\ }l}
r_1:& a \lor \tuple{\{a,b\},\{\{a\},\{a,b\}\}} & \\ r_2:& b \lor \tuple{\{a,b\},\{\{b\},\{a,b\}\}} & \end{array}
\]
\noindent
We have that $\{a,b\}$ is the only answer set of $\progCex{ex:unstableTDMa}$
and 
\[
\begin{array}{r@{}l}
C_1=&\tuple{\emptyset,\emptyset,\emptyset,\{\emptyset\}},\\&
    \tuple{\{r_1\},\{a,b\},\emptyset,\{\emptyset,\{b\}\}},\\&
    \tuple{\{r_1,r_2\},\{a,b\},\emptyset,\{\emptyset\}} \quad \quad \mbox{and}\\\\
C_2=&\tuple{\emptyset,\emptyset,\emptyset,\{\emptyset\}},\\&
    \tuple{\{r_2\},\{a,b\},\emptyset,\{\emptyset,\{a\}\}},\\&
    \tuple{\{r_1,r_2\},\{a,b\},\emptyset,\{\emptyset\}}
\end{array}
\]
\noindent
are the only computations that succeed for $\progCex{ex:unstableTDMa}$.
Clearly, $\progCex{ex:unstableTDMa}$ is monotone and absolutely tight but $C_1$ and $C_2$ are not stable.
\eex

Nevertheless, we can assure the existence of stable computations for answer sets of normal \cprograms that are absolutely tight and convex.
This is good news, as this class corresponds to a large subset of typical
answer-set programs written for solvers like \clasp that do not rely on
disjunction as their guessing device.

\begin{theorem}\label{th:unfoundedfreecomp}
Let $\computation\!=\!S_0,\dots,S_n$ be a computation
such that $S_0$ and $S_n$ are stable and
$\progC_\Delta=\stateprog{S_n}\setminus\stateprog{S_0}$ is a normal, convex, and absolutely tight \cprogram.
Then, there is a stable computation $\computation'\!=\!S'_0,\dots,S'_n$
such that
$S_0=S'_0$ and $S_n=S'_n$.
\end{theorem}
\noindent
As a direct consequence of Theorem~\ref{th:unfoundedfreecomp} and Corollary~\ref{cor:completeness},
we get an improved completeness result for normal convex \cprograms that are absolutely tight, \iec 
we can find a computation that consists of stable states only.
\begin{corollary}
Let $\progC$ be a normal \cprogram that is convex and absolutely tight, and consider some $I\in\AS{\progC}$.
Then, there is a rooted stable computation $S_0,\dots,S_n$ such that
$\stateprog{S_n}=\reductFLP{\progC}{\inter}$ and
$\stateI{S_n}=I$.
\end{corollary}
\bp
From $I\in\AS{\progC}$, we get by Corollary~\ref{cor:completeness} that 
there is a rooted computation
$S_0,\dots,S_n$ such that
$\stateprog{S_n}=\reductFLP{\progC}{\inter}$ and
$\stateI{S_n}=I$.
Note that $S_0$ is the empty state.
$S_0$ and $S_n$ are stable according to Definition~\ref{def:stable}.
From Theorem~\ref{th:unfoundedfreecomp}, we can conclude the existence of another computation
$\computation'\!=\!S'_0,\dots,S'_n$
such that
$S_0=S'_0$ and $S_n=S'_n$ that is stable.
Clearly, $\computation'$ is also rooted.
\ep

 \section{Theory and Practice of Stepping}\label{sec:stepping}
In this section we present our methodology for stepping answer-set programs
based on the computation model introduced in the previous section.

Step-by-step execution of a program is common practice in procedural programming languages,
where developers can debug and investigate the behaviour of their programs in an incremental way.
The technique introduced in this work
shows how this popular form of debugging can be applied to ASP,
despite the genuine declarative semantics of answer-set programs that lacks a control flow.
Its main application is debugging but it is also beneficial in other contexts such as
improving the understanding of a given answer-set program or
teaching the answer-set semantics to beginners.

For stepping to be a practical support technique for answer-set programmers rather than a purely theoretical approach,
we assume the \memph{availability of a support environment} that assists a user in
a stepping session. 
A prototype our stepping framework has been implemented in \sealion, an \memph{integrated development environment} (IDE) for ASP~\cite{OPT13}. 
It was developed as one of the major goals of 
a research project on methods and methodologies for developing answer-set programs conducted at Vienna University of Technology (2009-2013).
\sealion supports the ASP languages of \gringo and \dlv and comes as a plugin of the Eclipse platform that is popular for Java development.
All the features of a stepping support environment described in this section are implemented in \sealion, if not stated otherwise.

To bridge the gap between theory and practical stepping, our examples use solver syntax rather than the abstract language of c-programs.
We implicitly identify solver constructs with their abstract counterparts (\cf Section~\ref{sec:agexweightAsC}).
While we use c-programs as a formal lingua franca for different solver languages, we do not require the user to know about it.
Likewise, we do not expect a user to be aware of the specifics of the computation framework of Section~\ref{sec:computation} that provides the backbone of stepping.
For example, the user does not need to know the properties of Definition~\ref{def:successor}. The debugging environment automatically restricts the available choices a user has when performing a step to ones that result in valid successor states.

In the following subsection, we introduce an example scenario that we use later on.
In Section~\ref{sec:steppingIdea}, we describe the general idea of stepping for ASP.
There are two major ways for navigating in a computation in our framework:
performing \memph{steps}, discussed in Section~\ref{sec:steps},
and \memph{jumps} that we describe in Section~\ref{sec:jumping}.
In Section~\ref{sec:sealionStepping}, we describe the stepping interface of \sealion.
Building on steps and jumps, we discuss
methodological aspects of stepping for debugging purposes in Section~\ref{sec:methodology}
and provide a number of use cases.

\subsection{Example Scenario - Maze Generation}\label{sec:maze}
Next, we introduce the problem of \memph{maze generation}
that serves as a running example in the remainder of the paper.
It has been a benchmark problem 
of the second ASP competition~\citep{DVBGT09}
to which it was submitted by Martin Brain.
The original problem description is available on the competition's website.\footnote{\url{http://dtai.cs.kuleuven.be/events/ASP-competition/Benchmarks/MazeGeneration.shtml}}

As the name of the problem indicates,
the task we deal with is to generate a maze,
\iec a labyrinth structure in a
grid that satisfies certain conditions.
In particular, we deal with 
two-dimensional grids of cells
where each cell can be assigned to be either an empty space or a wall.
Moreover, there are two (distinct) empty squares on the edge of the grid, known as the entrance and the exit.
A path is a finite sequence of cells, in which each distinct cell appears at most once and each cell is horizontally or vertically adjacent to the next cell in the sequence. 

\begin{figure}[t]
{
\centering
\phantom{|}
\hfill
\includegraphics[width=0.3\textwidth]{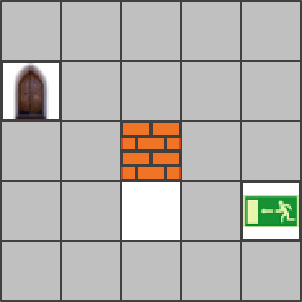}
\hfill
\includegraphics[width=0.3\textwidth]{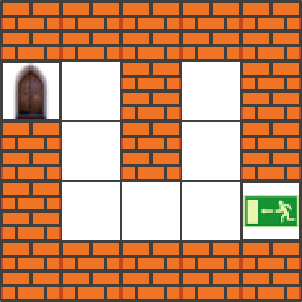}
\hfill
\phantom{|}
\caption{Left: A grid visualising an instance of the maze generation where white squares represent empty cells, whereas grey squares are yet undefined. Right: A solution for the instance on the left.}
\label{fig:mazeExample}

}
\end{figure}

Such a grid is a valid maze if it meets the following criteria:
\be
\item Each cell is a wall or is empty. 
\item There must be a path from the entrance to every empty cell (including the exit).
\item If a cell is on any of the edges of the grid, and is not an entrance or an exit, it must contain a wall.
\item There must be no 2x2 blocks of empty cells or walls.
\item No wall can be completely surrounded by empty cells.
\item If two walls are diagonally adjacent then one or other of their common neighbours must be a wall.
\ee
The maze generation problem is the problem
of completing a two-dimensional grid in which
some cells are already decided to be empty or walls 
and the entrance and the exit are pre-defined to a valid maze.
An example of a problem instance and a corresponding solution maze
is depicted in Fig.~\ref{fig:mazeExample}.

Next, we describe the predicate schema that we use for ASP maze generation encodings.
The predicates \verb|col/1| and \verb|row/1| define the columns and rows in the grid, respectively.
They are represented by a range of consecutive, ascending integers, starting at \verb|1|.
The positions of the entrance and the exit are determined by predicates 
\verb|entrance/2| and \verb|exit/2|, respectively, where the first argument is
a column index and the second argument is a row index.
In a similar manner, \verb|empty/2| and \verb|wall/2| determine which cells are empty or contain walls.
For example,
the instance of Fig.~\ref{fig:mazeExample} can be encoded by program \progLabel{exMazeInstance.gr} consisting of the following facts:
\begin{verbatim}
col(1..5). row(1..5).
entrance(1,2). exit(5,4). wall(3,3). empty(3,4).
\end{verbatim}
\noindent
Moreover, the solution in the figure could be represented by the following interpretation (projected to predicates \verb|empty/2| and \verb|wall/2|):
\begin{verbatim}
{wall(1,1), empty(1,2), wall(1,3), wall(1,4), wall(1,5),
 wall(2,1), empty(2,2), empty(2,3), empty(2,4), wall(2,5),
 wall(3,1), wall(3,2), wall(3,3), empty(3,4), wall(3,5),
 wall(4,1), empty(4,2), empty(4,3), empty(4,4), wall(4,5),
 wall(5,1), wall(5,2), wall(5,3), empty(5,4), wall(5,5)}
\end{verbatim}

\subsection{General Idea}\label{sec:steppingIdea}

We introduce stepping for ASP as a strategy to identify mismatches between the intended semantics of an answer-set program under development and its actual semantics.
Due to the declarativity of ASP, once one detects unintended semantics,
it can be a tough problem to manually detect the reason.
Stepping is a method for breaking this problem into smaller parts and
structuring the search for an error.
The general idea is to monotonically build up an interpretation by,
in each step, adding literals derived by a rule that is active
with respect to the interpretation obtained in the previous step.
The process is interactive in the sense that at each such step the user
chooses the active rule to proceed with and decides which
literals of the rule should be considered true or false in
the target interpretation.
Hereby, the user only adds rules he or she thinks are active in an expected or an unintended actual answer set.
The interpretation grows monotonically until it is eventually guaranteed to be an answer set of the overall program, otherwise the programmer is informed why and at which step something went wrong.
This way, one can in principle without any backtracking direct the computation towards the interpretation one has in mind.
In debugging, having the programmer in the role of an oracle is a common scenario
as it is reasonable to assume that a programmer has good intuitions on where to guide the search~\citep{S82}.
We use the computation model of Section~\ref{sec:computation} to ensure that,
if the interpretation
specified in this way is indeed an answer set,
the process of stepping will eventually terminate with the interpretation as its result.
Otherwise, the computation will fail at some step where the user gets insight why
the interpretation is not an answer set, \egc when a constraint becomes irrevocably active
or no further rule is active that could derive some desired literal.

\subsection{Steps}\label{sec:steps}
By a \memph{step} we mean the extension of a computation by a further state.
We consider a setting, where a programmer has written an answer-set program
in a solver language for which \cprogram \progC it the abstraction of its grounding.
Moreover, we assume that the programmer has obtained some computation for \progC
that is neither stuck in \progC nor complete for \progC.
For performing a step, one needs to find a successor state $S_{n+1}$ for $S_n$ such that $C' = S_0,\dots, S_{n+1}$ is a computation for \progC.

We propose 
a sequence of 
three user actions 
to perform a step.
Intuitively, for quickly finding a successor state (with the help of the debugging environment), we suggest to
\be
   \item select a non-ground rule with active ground instances, then
   \item choose an active ground rule among the instances of the non-ground rule, and
   \item select for yet undefined atoms in the domain of the ground instance whether they are considered true or false.
\ee
First, the user selects a non-ground rule $\ruleG$.
In \sealion, this can be done by directly selecting $\ruleG$ in the editor in which the program was written.
The debugging system can support this user action by
automatically
determining the subset of rules in the program that have at least one 
instance $r$ in their grounding that could lead to a successor state, \iec
$r=\newRule{S_n}{S}$ for some successor $S$ of $S_n$.

Then, the
user selects an instance $r$ from the grounding of $\ruleG$.
As the ground instances of $\ruleG$ are not part of the original program,
picking one requires a different approach as for choosing $\ruleG$.
Here, the debugging environment can display the ground rules in a dedicated area
and, as before, restrict the choice of rule groundings of \ruleG
to ones that lead to a successor state.
Filtering techniques can be used to restrict the amount of the remaining instances.
In \sealion, the user defines partial assignments for the variables in \ruleG
that determine a subset of the considered instances.

In the third user action for performing a step, the programmer
chooses the truth values for the atoms in $\domain{r}$ that are neither in
\stateI{S_n} nor in \statenegI{S_n}.
This choice must be made in a way such that
there is a successor $S_{n+1}$ of $S_{n}$
with 
$\stateprog{S_{n+1}}=\stateprog{S_{n}}\cup\{r\}$,
$\stateI{S_{n+1}}=\stateI{S_{n}}\cup\Delta$, and
$\statenegI{S_{n+1}}=\statenegI{S_{n}}\cup\negpart{\Delta}$,
where $\Delta$ contains the atoms the user chose to be true
and $\negpart{\Delta}$ the atoms considered false.
That is, $S_{n}$, $\Delta$, and $\negpart{\Delta}$ must fulfil the conditions of
Definition~\ref{def:successor}.
Here, the user needs only to ensure that
Condition~(v) of Definition~\ref{def:successor} holds, \iec
$\stateI{S_{n+1}}\models \body{r}$ and $\stateI{S_{n+1}}\modelsDisj\head{r}$,
as the other conditions automatically hold once all unassigned atoms have been assigned to $\Delta$ and $\negpart{\Delta}$.
In particular, the set of unfounded sets, 
$\stateunfounded{S_{n+1}}$ can always be automatically computed
following Condition~(vi) of Definition~\ref{def:successor}
and does not impose restrictions on the choice of $\Delta$ and $\negpart{\Delta}$.
The support system can check whether Condition~(v) holds for the truth assignment
specified by the user. Also, atoms are automatically assigning to 
$\Delta$ or $\negpart{\Delta}$ whenever their truth values are the same for all successor states that are based on adding $r$.

\bex{ex:step}
As a first step for developing the maze-generation encoding,
we want to identify border cells and guess an assignment of walls and empty cells.
Our initial program is
\progLabel{ex-ref-ex:step-.gr}, given next.

\begin{verbatim}
maxCol(X) :- col(X), not col(X+1).
maxRow(Y) :- row(Y), not row(Y+1).
border(1,Y) :- col(1), row(Y).
border(X,1) :- col(X), row(1).
border(X,Y) :- row(Y), maxCol(X).
border(X,Y) :- col(X), maxRow(Y).

wall(X,Y) :- border(X,Y), not entrance(X,Y), not exit(X,Y).
{ wall(X,Y) : col(X), row(Y), not border(X,Y) }.
empty(X,Y) :- col(X), row(Y), not wall(X,Y).
\end{verbatim}
\noindent
The first two rules extract the numbers of columns and rows of the maze from the input facts of predicates \verb|col/1| and \verb|row/1|.
The next four rules derive \verb|border/2| atoms that indicate which cells form the border of the grid. 
The final three rules 
 derive \verb|wall/2| atoms for border cells except entrance and exit,
 guess \verb|wall/2| atoms for the remaining cells,
 and derive \verb|empty/2| atoms for non-wall cells, respectively.

We use 
\progRef{ex-ref-ex:step-.gr}
in conjunction with the facts in program \progRef{exMazeInstance.gr} (defined in Section~\ref{sec:maze}) that determine the problem instance.

We start a stepping session with the computation $C_0=S_0$ consisting of the empty state $S_0=\tuple{\emptyset,\emptyset,\emptyset,\{\emptyset\}}$.
Following the scheme of user actions described above for performing a step,
we first look for a non-ground rule with instances that are active under $\stateI{S_0}$.
As $\stateI{S_0}=\emptyset$, only the facts from \progRef{exMazeInstance.gr} have active instances.
We choose the rule \verb|entrance(1,2).|
In this case, the only (active) instance of the rule is  
identical to the rule, \iec the fact:
\begin{verbatim}
entrance(1,2).
\end{verbatim}
The only atom in the domain of the rule instance is \verb|entrance(1,2)|.
Therefore, when performing the final user action for a step
one has to decide the truth value of this atom.
In order to fulfil Condition (v) of Definition~\ref{def:successor},
the rule head, \iec \verb|entrance(1,2)|, must be true in the successor state.
Thus, our first step results in the computation 
$C_1=S_0,S_1$ where 
\[
S_1=\langle\{\mathverb{entrance(1,2).}\},\{\mathverb{entrance(1,2)}\},\emptyset,\{\emptyset\}\rangle.
\]

For the next step, we choose the rule 
\begin{verbatim}
col(1..5).
\end{verbatim}
from \progRef{exMazeInstance.gr}.
The grounding\footnote{Remember that grounding refers to the translation performed by the grounding tool rather than mere variable elimination.} by \gringo consists of the following instances:
\begin{verbatim}
col(1). col(2). col(3). col(4). col(5).
\end{verbatim}
\noindent
We select the instance \verb|col(5).| 
Since the head of the rule must be true under the successor state,
as before, atom \verb|col(5)| must be considered true in the successor state of $S_1$.
The resulting computation after the second step is $C_2=S_0,S_1,S_2$, where 
\[
S_2=\langle\{\mathverb{entrance(1,2).} \mathverb{col(5).}\},\{\mathverb{entrance(1,2), \mathverb{col(5)}\}},\emptyset,\{\emptyset\}\rangle.
\]

Under $\stateI{S_2}$ a further rule in $\progRef{exMazeInstance.gr}\cup\progRef{ex-ref-ex:step-.gr}$
has active instances:
\begin{verbatim}
maxCol(X) :- col(X), not col(X+1).
\end{verbatim}
That is, it has the active instance
\begin{verbatim}
maxCol(5) :- col(5), not col(6).
\end{verbatim}
that we choose for the next step.
In order to ensure that Condition (v) of Definition~\ref{def:successor} is satisfied,
we need to ensure that head and body are satisfied under the successor state.
Hence, atom \verb|maxCol(5)| has to be considered true, whereas \verb|col(6)| must be considered false.
We obtain the computation $C_3=S_0,S_1,S_2,S_3$, where 
\[
\begin{array}{l@{}l@{}l}
S_3=\langle & \{
\mathverb{entrance(1,2).}
\mathverb{col(5).}
\mathverb{maxCol(5) :- col(5), not col(6).}\},\\&
\{\mathverb{entrance(1,2)}, \mathverb{col(5)}, \mathverb{maxCol(5)}\},\{\mathverb{col(6)}\},\{\emptyset\}\rangle.
\end{array}
\]
\end{example}

\subsection{Jumps}\label{sec:jumping}
If one wants to simulate the computation of an answer set $\inter$ in a stepping session using steps only,
as many steps are necessary as there are active rules in the grounding under $\inter$.
Although, typically the number of active ground instances is much less than the total number
of rules in the grounding, still many rules would have to be considered.
In order to focus on the parts of a computation that the 
user is interested in, we introduce a \memph{jumping} technique for quickly considering rules
that are of minor interest, \egc for rules that are already considered correct.
We say that we \memph{jump through} these rules.
By performing a jump, we mean to find a state that could be reached by a computation for the program at hand that extends the current computation by possibly multiple states.
If such a state can be found, one can continue to expand a computation from that while it is ensured that the same states could be reached by using steps only.
Jumps can be performed exploiting Theorem~\ref{th:jumping}.
In essence, jumping is done as follows.
\be
   \item Select rules that you want to jump through (\iec the rules you want to be considered in the state to jump to), 
   \item an auxiliary answer-set program is created that contains the selected rules and the active rules of the current computations final state, and
   \item a new state is computed from an answer set of the auxiliary program.
\ee
\noindent
Next, we describe the items in more detail.
We assume that an answer-set program in a solver language for which \cprogram \progC it the abstraction of its grounding
and a computation $S_0,\dots, S_{n}$ for \progC are given.

The first user action is to select a subset $\progC'$ of \progC, the rules to jump through.
Hence, an implication for a stepping support environment
is the necessity of means to select the ground instances that form $\progC'$.
In case the user wants to consider all instances of some non-ground rules, a very user friendly
way of selecting these is to simply select the non-ground rules and the environment implicitly considers $\progC'$ to be their instances.
\sealion implements this feature such that this selection can be done in the editor in which the answer-set program is written.
If the user wants to jump through only some instances of a non-ground rule,
we need further user-interface features.
In order to keep memory resources and the amount of rules that have to be considered by the user low, the system splits the selection of an instance in two phases.
First, the user selects a non-ground rule $\ruleG$, similar as in the first user action of defining a step.
Then, the system provides so far unconsidered rules of $\progC$ for selection,
where similar filtering techniques as sketched for the second user action for performing a step can be applied.

The auxiliary program of the second item can be automatically computed.
That is a \cprogram $\progC_{\mathit{aux}}$ given by
$\progC_{\mathit{aux}}=\stateprog{S_n}\cup\progC'\cup\progC_{\mathit{con}}$,
where 
\[
\begin{array}{l@{}l}
\progC_{\mathit{con}}=&\{\la \naf\ a\mid a\in \stateI{S_n}\}\ \cup\\
&\{\la a\mid a\in \statenegI{S_n}\}
\end{array}
\]
\noindent
is a set of constraints
that ensure that for every answer set $I$ of $\progC_{\mathit{aux}}$
we have $\stateI{S_n}\subseteq I$ and $I\cap\statenegI{S_n}=\emptyset$.

After computing an answer set $I$ of the auxiliary program,
Theorem~\ref{th:jumping}
ensures the existence of a computation
$C'=S_0,\dots,S_n,S_{n+1},\dots,S_m$ for $\progC$ such that 
$S_m$ is stable and $\stateI{S_m}=I$.
Moreover, we have $\stateprog{S_m}=\stateprog{S_n}\cup\reductFLP{\progC'}{\inter}$.
Then, the user can proceed with further steps or jumps extending
the computation $S_0,\dots,S_m$ as if $S_m$ had  been reached by steps only.

\begin{figure}[t]
{\centering
\includegraphics[width=\textwidth]{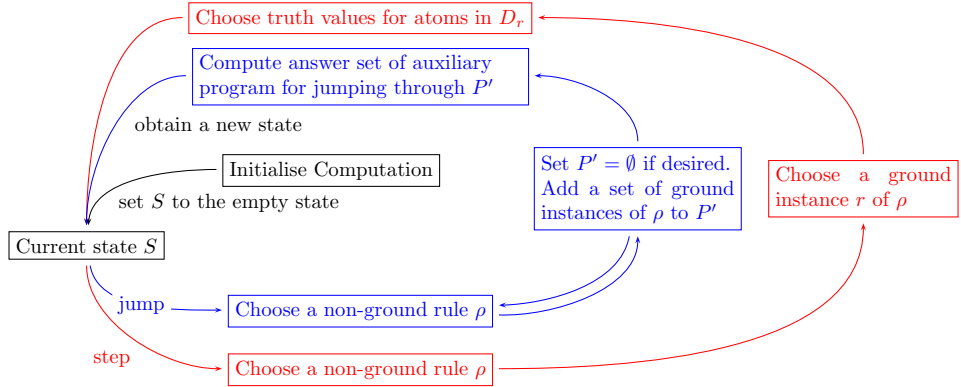}
\caption{Stepping cycle}
\label{fig:steppingCycle}
}
\end{figure}

Note that non-existence of answer sets of the auxiliary program
does not imply that the overall program has no answer sets as shown next.
\bex{ex:nojump}
Consider a program $\progC$ consisting of the \crules
$a \la$ and $\la\naf\ a$ that has $\{a\}$ as its unique answer set.
Assume we want to jump through the second rule
starting from the computation $C=\tuple{\emptyset,\emptyset,\emptyset,\{\emptyset\}}$ consisting of the empty state.
Then, $\progC_{\mathit{aux}}=\{\la\naf\ a\}$ has no answer set.
\eex
\noindent
The example shows that
jumping only makes sense when the user is interested
in a computation reaching an answer set of the auxiliary program.
In case of multiple answer sets of the auxiliary program,
the user could pick any or a stepping environment can choose one at random.
For practical reasons, the second option seems more preferable.
On the one hand, presenting multiple answer sets to the user can lead to a large amount of information
that has to be stored and processed by the user.
And on the other hand,
if the user is not happy with the truth value of some atoms in an arbitrary answer set
of the auxiliary program, he or she can use steps to define the truth of these atoms before performing the jump.
In \sealion only one answer set of the auxiliary program is computed.

The iterative extension of a computation using steps and jumps
can be described as a \memph{stepping cycle}
that is depicted in Fig.~\ref{fig:steppingCycle}.
It summarises how a user may advance a computation and thus provides a technical level representation of stepping.

\bex{ex:jump}
We continue computation $C_3$ for 
program $\progRef{exMazeInstance.gr} \cup\progRef{ex-ref-ex:step-.gr}$
from Example~\ref{ex:step}.
As we are interested in the final three rules of \progRef{ex-ref-ex:step-.gr}
that derive \verb|empty/2| and \verb|wall/2| atoms
but these rules depend on atoms of predicate \verb|border/2|, \verb|entrance/2|, and \verb|exit/2|
that are not yet considered in $C_3$,
we jump through the facts from \progRef{exMazeInstance.gr} and the rules 
\begin{verbatim}
maxCol(X) :- col(X), not col(X+1).
maxRow(Y) :- row(Y), not row(Y+1).
border(1,Y) :- col(1), row(Y).
border(X,1) :- col(X), row(1).
border(X,Y) :- row(Y), maxCol(X).
border(X,Y) :- col(X), maxRow(Y).
\end{verbatim}
of program \progRef{ex-ref-ex:step-.gr}.
The resulting auxiliary program \progLabel{ex-ref-ex:jump-aux.gr} is given by the following rules (for non-ground rules, their unconsidered instances in the grounding of $\progRef{exMazeInstance.gr} \cup\progRef{ex-ref-ex:step-.gr}$).
\begin{verbatim}
entrance(1,2).
col(5).
maxCol(5) :- col(5), not col(6).

col(1..5). row(1..5).
exit(5,4). wall(3,3). empty(3,4).
maxCol(X) :- col(X), not col(X+1).
maxRow(Y) :- row(Y), not row(Y+1).
border(1,Y) :- col(1), row(Y).
border(X,1) :- col(X), row(1).
border(X,Y) :- row(Y), maxCol(X).
border(X,Y) :- col(X), maxRow(Y).

:- not entrance(1,2).
:- not col(5).
:- not maxCol(5).
:- col(6).
\end{verbatim}
\noindent
The program \progRef{ex-ref-ex:jump-aux.gr}
has the single answer set $\inter_{\mathit{aux}}$ consisting of the atoms:
\begin{verbatim}
col(1), col(2), col(3), col(4), col(5), maxCol(5), 
row(1), row(2), row(3), row(4), row(5), maxRow(5),
empty(3,4), wall(3,3), entrance(1,2), exit(5,4),
border(1,1), border(2,1), border(3,1), border(4,1),
border(5,1), border(1,2), border(5,2), border(1,3),
border(5,3), border(1,4), border(5,4), border(1,5),
border(2,5), border(3,5), border(4,5), border(5,5),
\end{verbatim}
\noindent
We obtain the new state $S_4=\tuple{\stateprog{S_4},\inter_{\mathit{aux}},\domain{\stateprog{S_4}}\setminus \inter_{\mathit{aux}},\{\emptyset\}}$,
where $\stateprog{S_4}$ consists of the following rules:
\begin{verbatim}
col(1). col(2). col(3). col(4). col(5).
row(1). row(2). row(3). row(4). row(5).
wall(3,3). empty(3,4). entrance(1,2). exit(5,4). 
maxCol(5) :- col(5), not col(6).
maxRow(5) :- row(5), not row(6).
border(1,1) :- col(1), row(1).
border(2,1) :- col(2), row(1).
border(3,1) :- col(3), row(1).
border(4,1) :- col(4), row(1).
border(5,1) :- col(5), row(1).
border(1,2) :- col(1), row(2).
border(5,2) :- row(2), maxCol(5).
border(1,3) :- col(1), row(3).
border(5,3) :- row(3), maxCol(5).
border(1,4) :- col(1), row(4).
border(5,4) :- row(4), maxCol(5).
border(1,5) :- col(1), row(5).
border(5,1) :- row(1), maxCol(5).
border(5,5) :- row(5), maxCol(5).
border(1,5) :- col(1), maxRow(5).
border(2,5) :- col(2), maxRow(5).
border(3,5) :- col(3), maxRow(5).
border(4,5) :- col(4), maxRow(5).
border(5,5) :- col(5), maxRow(5).
\end{verbatim}
Theorem~\ref{th:jumping} ensures 
the existence of a computation 
$C_4=S_0,S_1,S_2,$ $S_3,\dots,S_4$
for program $\progRef{exMazeInstance.gr}\cup\progRef{ex-ref-ex:step-.gr}$.
\eex

\subsection{Stepping Interface of \sealion}\label{sec:sealionStepping}

In the following, we focus on the stepping functionality of \sealion that was implemented by
our former student Peter Sko\v{c}ovsk\'{y}.
While it is the first implementation of the stepping technique for ASP and
hence still a prototype, it is tailored for intuitive and user-friendly usage and able to cope with real-world answer-set programs.
The stepping feature is integrated with the 
\kara plugin of \sealion~\cite{KOPT13}
that can create user-defined graphical representations of interpretations.
Thus, besides visualising answer sets, it is also possible to visualise intermediate states of a stepping session.
Visualisations in \kara are defined using ASP itself, for further information we refer to earlier work~\cite{KOPT13}.
A comprehensive discussion of other features of \sealion is given in a related paper~\citep{BOPST13} on the IDE.
\sealion is published under the GNU General Public License
and can be obtained from
\url{www.sealion.at}

A stepping session in \sealion can be started in a similar fashion
as debugging Java programs in Eclipse using the launch configuration framework.
\sealion launch configurations that are used for defining which program files should
be run with which solvers can be re-used
as \memph{debug configurations}.

Like many IDEs, Eclipse comes with a multiple document interface in which inner frames, in particular Eclipse editors and views, can 
be arranged freely by the user.
Such configurations can be persisted as \memph{perspectives}.
Eclipse plugins often come with default perspectives, \iec arrangements of views and editors that
are tailored to a specific user task in the context of the plugin.
Also the stepping plugin has a preconfigured perspective
that is opened automatically once a stepping session has been initiated.
The next subsection gives an overview of the individual stepping related subframes in this perspective.

\begin{figure}[t!]
{\centering
 \includegraphics[width=1\textwidth]{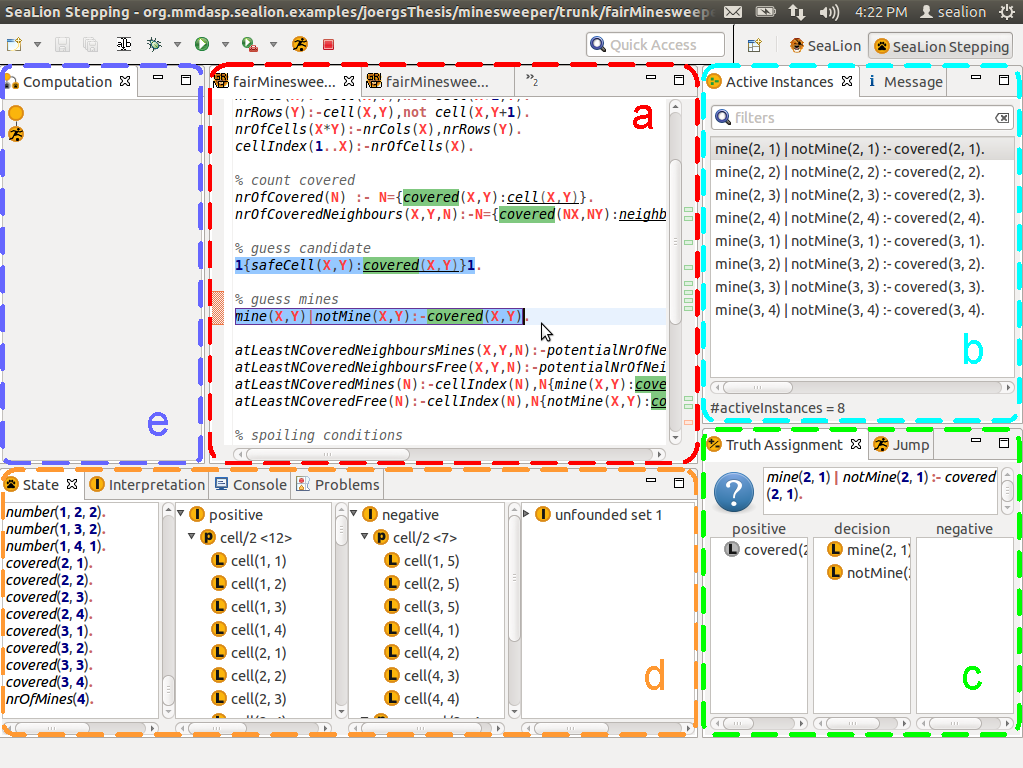}
\caption{\sealion's stepping view is devided into five areas (a-e).}
\label{fig:steppingPerspective}

}
\end{figure}

Fig.~\ref{fig:steppingPerspective} shows \sealion in the \memph{stepping perspective}.
The illustration distinguishes five regions (marked by supplementary dashed frames and labelled by letters)
for which we give an overview in what follows.

\begin{figure}[t]
{
\centering
 \includegraphics[width=1\textwidth]{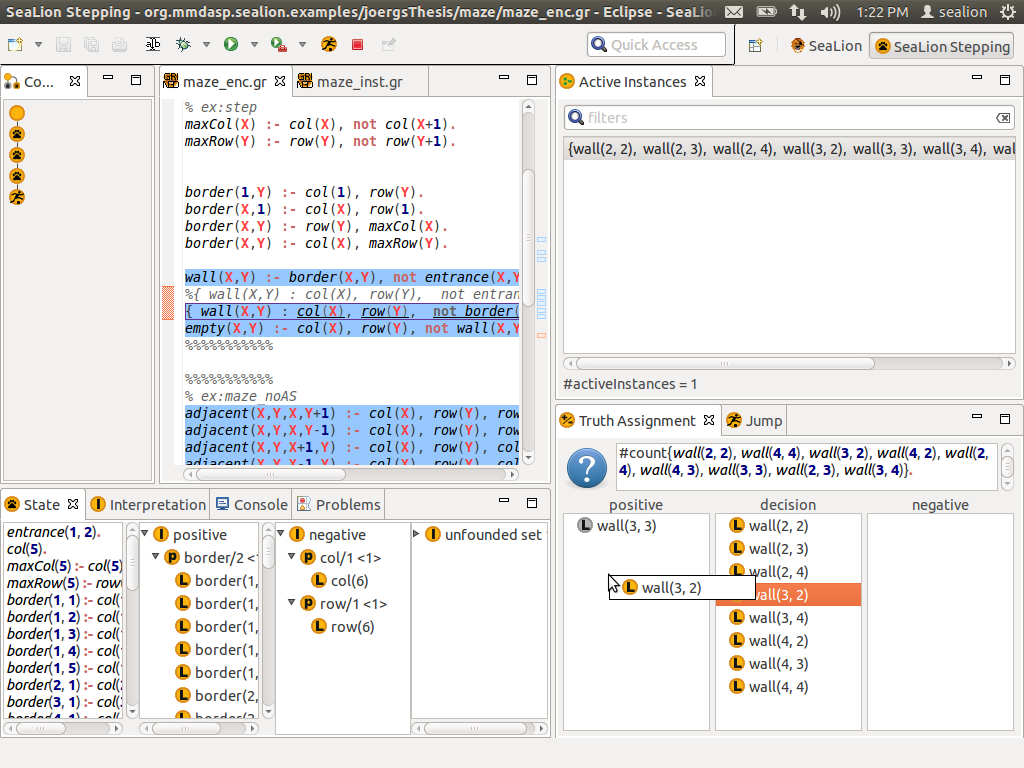}
\caption{
Deciding to consider atom 
\texttt{wall(3,2)}
to be true by dragging it from the middle list of atoms in the truth assignment view and dropping it in the left list.
The current state of the stepping session is $S_4$ from Example~\ref{ex:maze_AS} and the step prepared is that to state $S_5$.
Note that atom 
\texttt{wall(3,3)}
is greyed in the list of positive atoms.
Greyed atoms in the positive or negative list cannot be be dragged away from there again because their truth value is already considered 
positive, respectively negative, in the current state (here $S_4$).
A step can only be completed once the truth value has been decided for all atoms in the rule instance.
}
\label{fig:dragNdropTruthAssignment}

}
\end{figure}

The source code editor (Fig.~\ref{fig:steppingPerspective}a) is 
the same as used for writing answer-set programs
but extended with additional functionality during stepping mode for the ASP files involved in the stepping session.
In particular it indicates rules with ground instances that are active under the interpretation of the current stepping state.
Constraints with active instances are highlighted by a red background (\cf\ Fig.~\ref{fig:stuckConstraint}), other rules with active instances
have a blue background (as, \egc in Fig.~\ref{fig:dragNdropTruthAssignment}).
The editor remains functional during stepping, \iec the program can be modified while debugging.
Note, however, that the system does not guarantee that the current computation is still a valid computation
in case of a modification of the answer-set program after stepping has been initiated.
The source code editor is also the starting point for performing a step or a jump
as it allows for directly selecting the non-ground rule(s) to be considered in the step or jump in the source code.
The choice of non-ground rules corresponds to 
the initial step in the stepping cycle (see Section~\ref{sec:jumping}).
Selecting a single rule or consecutive rules is done by directly selecting them in the source code editor.
If the rules are non-consecutive, the user must collect rules in the \memph{jump view} located in area c of Fig.~\ref{fig:steppingPerspective} as the second tab.

Choosing a ground instance for performing a step is done in the \memph{active instances view} (Fig.~\ref{fig:steppingPerspective}b).
It contains a list with all active ground instances (with respect to conditional grounding) of the currently selected rule in the source editor.
As these are potentially many, the view has a textfield for filtering the list of rules.
Filters are given as dot-separated list of variable assignments of the form \texttt{X=t} where \texttt{X} is a variable of the non-ground rule
and \texttt{t} is the ground term that the user considers \texttt{X} to be assigned to.
Only ground instances are listed that obey all variable substitutions of the entered filters.

Once a rule instance is selected in the active instances view
the atoms in the rule's domain are displayed in three lists of the \memph{truth assignment view} (Fig.~\ref{fig:steppingPerspective}c).
The list in the centre shows atoms whose truth value has not already been determined in the current state.
The user can decide whether they should be true, respectively false, in the next step by putting them into the list on the left, respectively, on the right.
These atoms can be transferred between the lists by using keyboard cursors or drag-and-drop (Fig.~\ref{fig:dragNdropTruthAssignment}).
After the truth value has been decided for all the atoms of the rule instance and only
in case that the truth assignment leads to a valid successor state (\cf\ Definition~\ref{def:successor}),
a button labelled ``Step'' appears. Clicking this button computes the new state.

The state view (Fig.~\ref{fig:steppingPerspective}d) shows the current stepping state of the debugging session.
Hence, it is updated after every step or jump.
It comprises four areas, corresponding to the components of the state (\cf\ Definition~\ref{def:state}),
the list of active rules instances, 
a tree-shaped representation of the atoms considered true, 
a tree-shaped representation of the atoms considered false, 
both in a similar graphical representation as that of interpretations in the interpretation view,
and an area displaying the unfounded sets in a similar way.
The sets of atoms displayed in this view can also be visualised using \kara (via options in the context menu).

\begin{figure}[t]
{\centering
 \includegraphics[width=1\textwidth]{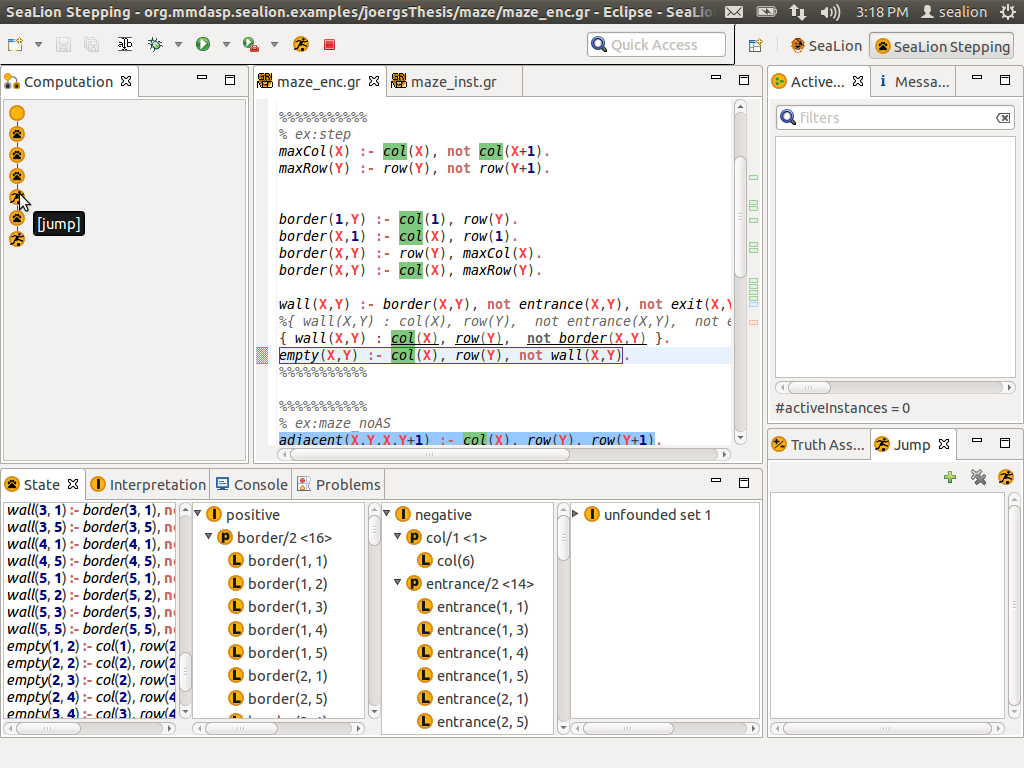}
}
\caption{
In \sealion, we can retract the computation to a previous state by clicking on the node in the computation view representing the step or jump that created the target state.
The screenshot shows reverting the last two states in Example~\ref{ex:mazeUnderstanding},
where we reused a part of the computation $S_1,\dots,S_6$ for exploring an alternative setting starting from $S_4$. The final state of the alternative computation presented in the example is depicted in Fig.~\ref{fig:stuckConstraint}.}
\label{fig:computationView}
\end{figure}

\begin{figure}[t]
{\centering
 \includegraphics[width=1.00\textwidth]{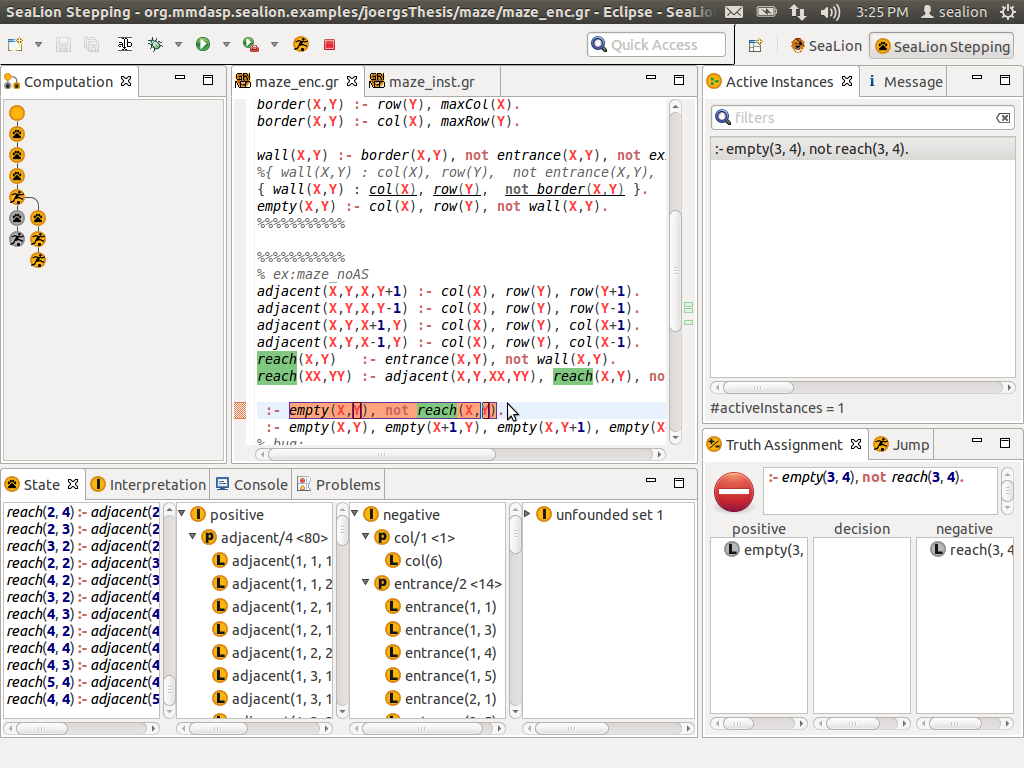}
\caption{
  The computation is stuck is state $S'_7$ from Example~\ref{ex:mazeUnderstanding} as only a constraint has active instances (highlighted in red in the editor). The ``No entry'' symbol in the truth assignment view indicates an unsatisfied instance under the current state.
  Note that the tree in the computation view has an inactive branch.
  The current computation (nodes with yellow background) is an alternative branch
  for the computation in Fig.~\ref{fig:computationView}. Clicking on the final greyed branch would set $S_1,\dots,S_6$ to be the active computation again.
}
\label{fig:stuckConstraint}

}
\end{figure}

Finally, the computation view (Fig.~\ref{fig:steppingPerspective}e) gives an overview of the steps and jumps performed so far.
Importantly, the view implements an undo-redo mechanism.
That is, by clicking on one of the nodes displayed in the view,
representing a previous step or jump, the computation can be reset to the state after this step or jump has been performed.
Moreover, after performing an undo operation, the undone computation is not lost but becomes an inactive branch of the
tree-shaped representation of steps and jumps.
Thus, one can immediately jump back to any state that has been reached in the stepping session by clicking on a respective node in the tree (Fig.~\ref{fig:computationView}).

Mismatches between the users intentions (reflected in the current stepping state) and the actual semantics of the program
can be detected in different parts of the stepping perspective.
If the user thinks a rule instance should be active but it is not,
this can already be seen in the source code editor if the non-ground version of the rule does not have any active instance.
Then, the rule is not highlighted in the editor.
If the non-ground version does have active instances but not the one the user has in mind,
this can be detected after clicking on the non-ground rule if they are missing in the active instances view.
The computation is stuck if only rules are highlighted in the source editor that are constraints (\cf\ Fig.~\ref{fig:stuckConstraint})
or for all of its instances, no truth assignment can be established such that the ``Step'' button appears.

Finally, if no further rule is highlighted and there is no non-empty unfounded set visible in the state view,
the atoms considered positive form an answer set of the overall program.
If there are further unfounded sets, the user sees that the constructed interpretation is not stable.
The unfounded sets indicate which atoms would need external support.

\subsection{Methodology}\label{sec:methodology}

We identify three three conceptual levels of stepping as a methodology.
The \memph{technical level} corresponds to the iterative advancement of a computation covered in Sections~\ref{sec:steps} and \ref{sec:jumping} summarised in the \memph{stepping cycle} (Fig.~\ref{fig:steppingCycle}).
Next, we describe stepping on the \memph{application level} as a method for \memph{debugging} and \memph{program analysis}.
After that, we highlight how stepping is \memph{embedded in} the greater context of \memph{ASP development} from a \memph{top level} perspective.
Finally, we illustrate our approach in different usage scenarios.
In \ifCORR \ref{sec:guidelines}\else supplementary Appendix A\fi, we compile practical guidelines for our methodology.

\subsubsection*{Program Analysis and Debugging Level Methodology}The main purpose for stepping in the context of this paper is its
application for debugging and analysing answer-set programs.
Next, we describe how insight into a program is gained using stepping.
During stepping, the user follows 
his or her intuitions on which rule(s) to apply next and
which atoms to consider true or false.
In this way, an interpretation is built up
that either is or is not an answer set of the program.
In both cases, stepping can be used to analyse the interplay of rules in the program in the same manner, \iec
one can see which rule instances become active or inactive after each step or jump.
In the case that the targeted interpretation is an answer set of the program,
the computation will never fail (in the sense of Definition~\ref{def:computationsemantics2}) or get stuck and will finally succeed.
It can, however, happen that intermediate states in the computation are unstable (\cf\ Example~\ref{ex:unstableHCNM}).
For debugging, stepping towards an answer set is useful if the answer set is unwanted.
In particular, one can see why a constraint (or a group of rules supposed to have a constraining effect)
does not become active.
For instance, stepping reveals other active rules that derive atoms that make some literal in the constraint false
or rules that fail do derive atoms that activate the constraint.
Stepping towards an actual answer set of a program is illustrated in Example~\ref{ex:maze_AS}.

In the case that there is an answer set that the user expects to be different, \iec certain atoms
are missing or unwanted, it makes sense to follow the approach that we recommend for expected but missing answer sets, \iec
stepping towards the interpretation that the user wants to be an answer set.
Then, the computation is guaranteed to fail at some point, \iec
there is some state in the computation from which no more answer set of the program can be reached.
In other situations, the computation can already have failed before
the bug can be found, \egc the computation can have failed from the beginning in case the program has no answer sets at all.
Nevertheless, the error can be found when stepping towards the intended interpretation.
In most cases, there will be either a rule instance that becomes active that the user considered inactive, or the other way around, \iec
a rule instance never becomes active or is deactivated while the computation progresses.
Eventually, due to our completeness results in the previous section, the computation will either get stuck or
ends in an unstable state $S$ such that no active external support for a non-empty unfounded set from $\stateunfounded{S}$
is available in the program's grounding.
Stepping towards an interpretation that is not an answer set of the overall program can be
seen as a form of hypothetical reasoning: the user can investigate how
rules of a part of the program support each other before adding a further rule would
cause an inconsistency.
Example~\ref{ex:mazeNoAS} illustrates stepping towards an intended but non-existing answer set for finding a bug.
Another illustration of hypothetical reasoning is given in Example~\ref{ex:mazeUnderstanding} where a user tries to
understand why an interpretation that is not supposed to be an answer set is indeed no answer set.

 Note that stepping does not provide explanation artefacts, analogous to stepping in imperative languages, where insight in a problem is gained by simply following the control flow and watching variable assignments.
 In our setting, the user only sees which remaining rule instances are active and the current state of the computation which in principle suffices to notice differences to his or her expectations. Nevertheless, it can be useful to combine this approach with query based debugging methods that provide explicit explanations as will be discussed in Section~\ref{sec:related}.
 
As mentioned in Section~\ref{sec:steppingIdea},
if one has a clear idea on the interpretation one expects to be an answer set,
stepping allows for building up a computation for this interpretation without backtracking.
In practice, one often lacks a clear vision on the truth value of each and every atom with respect to a desired answer set.
As a consequence, the user may require revising the decisions he or she has taken 
on the truth values of atoms as well as on which rules to add to the computation.
For this reason, \sealion allows for retracting a computation to a previous state, \iec
let the user select one of the states in the computation and continue stepping from there.
This way a tree of states can be built up (as in Fig.~\ref{fig:stuckConstraint}),
where every path from the root node to a leaf node is a rooted computation.

\subsubsection*{Top-Level Methodology}Stepping must be understood as embedded in the programming and modelling process, \iec
the technique has to be recognised in the \memph{context} of developing answer-set programs.
A practical consequence of viewing stepping in the big picture
are several possibilities for exploiting information obtained during the development
of a program for doing stepping faster and more accurate.

While an answer-set program evolves,
the programmer will in many cases
compute answer sets of preliminary versions of the program
for testing purposes.
If this answer sets are persisted, they can often be used as a starting point for 
stepping sessions for later versions of the program.
For instance, in case the grounding $\progC$ of a previous version of a program
is a subset of the current grounding $\prog'$ it is obvious
that a successful computation $C$ for $\progC$ is also a computation for $\progC'$.
Hence, the user can initiate a stepping session starting from $C$.
Also in case that $\progC\not\subset\progC'$,
a stepping support system could often
automatically build a computation that uses an answer set of $\progC$ (or parts of it)
as guidance for deciding on the rules to add and the truth values to assign in consecutive states.
Likewise, (parts of) computations of stepping sessions for previous versions of a program can be stored and re-used either as a computation of the current program if applicable or for computing such a computation.
The idea is that previous versions of a program often constitute a part of the current
version that one is already familiar with and ``trusts'' in.
By starting from a computation that already considers a well-known part of the program,
the user can concentrate on new and often more suspicious parts of the program.
Currently, there is no such feature implemented in \sealion.

\subsubsection*{Use Cases}Next, we show application scenarios of stepping using our running example.

The first scenario illustrates stepping towards an interpretation that is an answer set of the program under consideration.
\bex{ex:maze_AS}
We want to step towards an answer set of our partial encoding of the maze generation problem, \iec of the program $\progRef{exMazeInstance.gr}\cup\progRef{ex-ref-ex:step-.gr}$.
Therefore, we continue our stepping session with computation $C_4$, \iec we start stepping from state $S_4$ that we obtained in Example~\ref{ex:jump}.
In particular, we want to reach an answer set that is compatible with the maze generation solution
depicted in Fig.~\ref{fig:mazeExample}.
To this end,
we start with stepping through the active instances of the rule 
\begin{verbatim}
{wall(X,Y) : col(X), row(Y), not border(X,Y)}.
\end{verbatim}
The only active instance of the rule is
\begin{verbatim}
{wall(2,2), wall(3,2), wall(4,2), wall(2,3), wall(3,3),
            wall(4,3), wall(2,4), wall(3,4), wall(4,4)}.
\end{verbatim}
Thus, we next choose a truth assignment for the atoms appearing in the instance's choice atom.
Note that we do not need to decide for the truth value of \verb|wall(3,3)| as it is already contained in $\stateI{S_4}$ and therefore already considered true.
As can be observed in Fig.~\ref{fig:mazeExample}, among the remaining cells the rule deals with, only the one at position $(3,2)$ is
a wall in our example.
Hence, we obtain a new state $S_5$ from $S_4$ by extending $\stateprog{S_4}$ by our rule instance,
$\stateI{S_4}$ by \verb|wall(3,2)|, and $\statenegI{S_4}$ by \verb|wall(2,2)|, \verb|wall(4,2)|, \verb|wall(2,3)|, \verb|wall(4,3)|, \verb|wall(2,4)|, \verb|wall(3,4)|, \verb|wall(4,4)|.
As in $S_4$, the empty set is the only unfounded set in state $S_5$.
It remains to jump through the rules
\begin{verbatim}
wall(X,Y) :- border(X,Y), not entrance(X,Y), not exit(X,Y).
\end{verbatim}
\noindent
and
\begin{verbatim}
empty(X,Y) :- col(X), row(Y), not wall(X,Y).
\end{verbatim}
\noindent
that leads to the addition of instances
\begin{verbatim}
wall(1, 1) :- border(1, 1), not entrance(1, 1), not exit(1, 1).
wall(2, 1) :- border(2, 1), not entrance(2, 1), not exit(2, 1).
wall(3, 1) :- border(3, 1), not entrance(3, 1), not exit(3, 1).
wall(4, 1) :- border(4, 1), not entrance(4, 1), not exit(4, 1).
wall(5, 1) :- border(5, 1), not entrance(5, 1), not exit(5, 1).
wall(5, 2) :- border(5, 2), not entrance(5, 2), not exit(5, 2).
wall(1, 3) :- border(1, 3), not entrance(1, 3), not exit(1, 3).
wall(5, 3) :- border(5, 3), not entrance(5, 3), not exit(5, 3).
wall(1, 4) :- border(1, 4), not entrance(1, 4), not exit(1, 4).
wall(1, 5) :- border(1, 5), not entrance(1, 5), not exit(1, 5).
wall(2, 5) :- border(2, 5), not entrance(2, 5), not exit(2, 5).
wall(3, 5) :- border(3, 5), not entrance(3, 5), not exit(3, 5).
wall(4, 5) :- border(4, 5), not entrance(4, 5), not exit(4, 5).
wall(5, 5) :- border(5, 5), not entrance(5, 5), not exit(5, 5).
empty(1, 2) :- col(1), row(2), not wall(1, 2).
empty(2, 2) :- col(2), row(2), not wall(2, 2).
empty(4, 2) :- col(4), row(2), not wall(4, 2).
empty(2, 3) :- col(2), row(3), not wall(2, 3).
empty(4, 3) :- col(4), row(3), not wall(4, 3).
empty(2, 4) :- col(2), row(4), not wall(2, 4).
empty(3, 4) :- col(3), row(4), not wall(3, 4).
empty(4, 4) :- col(4), row(4), not wall(4, 4).
empty(5, 4) :- col(5), row(4), not wall(5, 4).	
\end{verbatim}
to a new state $S_6$.
$\stateI{S_6}$ extends $\stateI{S_5}$ by the head atoms of these rules that are not yet in $\stateI{S_5}$.
Likewise, 
$\statenegI{S_6}$ extends $\statenegI{S_5}$ by the default negated atoms appearing in the rules that are not yet in $\statenegI{S_5}$.
As $\stateunfounded{S_6}=\{\emptyset\}$ and no rule in $\progRef{exMazeInstance.gr}\cup\progRef{ex-ref-ex:step-.gr}$ has further active instances under
$\stateI{S_6}$, the computation $S_0,\dots,S_6$ has succeeded and hence $\stateI{S_6}$ is an answer set of the program.
\eex

In the next example, a bug is revealed by stepping towards an intended answer~set.
\bex{ex:mazeNoAS}
As a next feature,
we (incorrectly) implement rules in program~\progLabel{ex-ref-ex:mazeNoAS-.gr} that should express that there has to be a path from the entrance to every empty cell and that $2\times 2$ blocks of empty cells are forbidden:
\begin{verbatim}
adjacent(X,Y,X,Y+1) :- col(X), row(Y), row(Y+1).
adjacent(X,Y,X,Y-1) :- col(X), row(Y), row(Y-1).
adjacent(X,Y,X+1,Y) :- col(X), row(Y), col(X+1).
adjacent(X,Y,X-1,Y) :- col(X), row(Y), col(X-1).
reach(X,Y)   :- entrance(X,Y), not wall(X,Y).
reach(XX,YY) :- adjacent(X,Y,XX,YY), reach(X,Y), not wall(XX,YY).

:- empty(X,Y), not reach(X,Y).
:- empty(X,Y), empty(X+1,Y), empty(X,X+1), empty(X+1,Y+1).
\end{verbatim}
\noindent
The first six rules formalise when an empty cell is reached from the entrance, and the two constraints
should ensure that every empty cell is reached and that 
no $2\times 2$ blocks of empty cells exist, respectively.

Assume that we did not spot the bug in the second constraint---in the third body literal the term \verb|Y+1| was mistaken for \verb|X+1|.
This could be the result of a typical copy-paste error.
It turns out that $\progRef{exMazeInstance.gr}\cup\progRef{ex-ref-ex:step-.gr}\cup\progRef{ex-ref-ex:mazeNoAS-.gr}$ has no answer set.
In order to find a reason, one can start stepping towards an intended answer set.
We assume that the user already trusts 
the program $\progRef{exMazeInstance.gr}\cup\progRef{ex-ref-ex:step-.gr}$ from Example~\ref{ex:maze_AS}.
Hence, he or she can
reuse the computation $S_0,\dots,S_6$ 
for $\progRef{exMazeInstance.gr}\cup\progRef{ex-ref-ex:step-.gr}$
as starting point for a stepping session 
because all rules in $\stateprog{S_6}$ are also ground instances of rules in the extended program
$\progRef{exMazeInstance.gr}\cup\progRef{ex-ref-ex:step-.gr}\cup\progRef{ex-ref-ex:mazeNoAS-.gr}$.
Then, when the user asks for rules with active ground instances
a stepping support environment would present the following rules:
\begin{verbatim}
adjacent(X,Y,X,Y+1) :- col(X), row(Y), row(Y+1).
adjacent(X,Y,X,Y-1) :- col(X), row(Y), row(Y-1).
adjacent(X,Y,X+1,Y) :- col(X), row(Y), col(X+1).
adjacent(X,Y,X-1,Y) :- col(X), row(Y), col(X-1).
reach(X,Y)   :- entrance(X,Y), not wall(X,Y).

:- empty(X,Y), not reach(X,Y).
:- empty(X,Y), empty(X+1,Y), empty(X,X+1), empty(X+1,Y+1).
\end{verbatim}
The attentive observer will immediately notice that two
constraints are currently active.
\begin{figure}[t!]
\centering
\setlength{\columnsep}{38pt}
\begin{footnotesize}
\begin{multicols}{2}
\begin{verbatim}
col(1..5). row(1..5).
entrance(1,2). exit(5,4).
wall(3,3). empty(3,4).

maxCol(X) :- col(X), not col(X+1).
maxRow(Y) :- row(Y), not row(Y+1).
border(1,Y) :- col(1), row(Y).
border(X,1) :- col(X), row(1).
border(X,Y) :- row(Y), maxCol(X).
border(X,Y) :- col(X), maxRow(Y).

wall(X,Y) :- border(X,Y), 
             not entrance(X,Y),
             not exit(X,Y).
{ wall(X,Y) : col(X), row(Y),
              not border(X,Y) }.
empty(X,Y) :- col(X), row(Y),
              not wall(X,Y).

adjacent(X,Y,X,Y+1) :- col(X), row(Y),
                       row(Y+1).
adjacent(X,Y,X,Y-1) :- col(X), row(Y),
                       row(Y-1).
adjacent(X,Y,X+1,Y) :- col(X), row(Y),
                       col(X+1).
adjacent(X,Y,X-1,Y) :- col(X), row(Y),
                       col(X-1).
reach(X,Y)   :- entrance(X,Y), 
                not wall(X,Y).
reach(XX,YY) :- adjacent(X,Y,XX,YY),
                reach(X,Y),
                not wall(XX,YY).

:- empty(X,Y), not reach(X,Y).
:- empty(X,Y), empty(X+1,Y),
   empty(X,Y+1), empty(X+1,Y+1).

:- exit(X,Y), wall(X,Y).
:- wall(X,Y), wall(X+1,Y),
   wall(X,Y+1), wall(X+1,Y+1).
:- wall(X,Y), empty(X+1;X-1,Y),
   empty(X,Y+1;Y-1), col(X+1;X-1),
   row(Y+1;Y-1).
:- wall(X,Y), wall(X+1,Y+1), 
   not wall(X+1,Y), not wall(X,Y+1).
:- wall(X+1,Y), wall(X,Y+1), 
   not wall(X,Y), not wall(X+1,Y+1).
\end{verbatim} 
\end{multicols}
\end{footnotesize}
\caption{Program $\progRef{exMazeInstance.gr}\cup\progRef{ex-ref-ex:step-.gr}\cup\progRef{ex-ref-ex:mazeNoAS-b.gr}\cup\progRef{ex-ref-ex:mazeUnderstanding-.gr}$ used in Example~\ref{ex:mazeUnderstanding}.}
\label{fig:encodingReminder}
\end{figure}
There is no reason to be deeply concerned about
\begin{verbatim}
:- empty(X,Y), not reach(X,Y).
\end{verbatim}
being active because the rule defining the \verb|reach/2| predicate---that can potentially deactivate the constraint---has not been considered yet.
However, the constraint
\begin{verbatim}
:- empty(X,Y), empty(X+1,Y), empty(X,X+1), empty(X+1,Y+1).
\end{verbatim}
contains only atoms of predicate \verb|empty/2| that has already been fully evaluated.
Even if \verb|empty/2| was only partially evaluated,
an active instance of the constraint could not become inactive in a subsequent computation
for the sole ground that it only contains monotonic literals.
When the user inspects the single ground instance 
\begin{verbatim}
:- empty(1,2), empty(2,2), empty(1,2), empty(2,3).
\end{verbatim}
\noindent
of the constraint the bug becomes obvious.
A less attentive observer would maybe not immediately realise that
the constraint will not become inactive again.
In this case, he or she would in the worst case step through all the other rules
before the constraint above remains as the last rule with active instances.
Then, at the latest, one comes to the same conclusion that \verb|X+1| has to be replaced by \verb|Y+1|.
Moreover, a stepping environment could give a warning when there is a constraint instance that is guaranteed to stay active in subsequent states.
This feature is not implemented in \sealion.
We refer to the corrected version of program \progRef{ex-ref-ex:mazeNoAS-.gr}
by \progLabel{ex-ref-ex:mazeNoAS-b.gr}.
\eex

Compared to traditional software, programs in ASP are typically very succinct and often authored by a single person.
Nevertheless, people are sometimes confronted with ASP code written by another person, \egc in case of
joint program development, software quality inspection, legacy code maintenance, or evaluation of student assignments in a logic-programming course.
As answer-set programs can model complex problems within a few lines of code,
it can be pretty puzzling to understand someone else's ASP code, even if the program is short.
Here, stepping can be very helpful to get insight into how a program works that was written by another programmer,
as illustrated by the following example.

\bex{ex:mazeUnderstanding}
Assume that the full encoding of the maze generation encoding is composed by the programs 
$\progRef{ex-ref-ex:step-.gr}\cup\progRef{ex-ref-ex:mazeNoAS-b.gr}$
and that the constraints in \progLabel{ex-ref-ex:mazeUnderstanding-.gr}, given next, has been written by another author.
\begin{verbatim}
:- exit(X,Y), wall(X,Y).
:- wall(X,Y), wall(X+1,Y), wall(X,Y+1), wall(X+1,Y+1).
:- wall(X,Y), empty(X+1;X-1,Y), empty(X,Y+1;Y-1), col(X+1;X-1),
                                                  row(Y+1;Y-1).
:- wall(X,Y), wall(X+1,Y+1), not wall(X+1,Y), not wall(X,Y+1).
:- wall(X+1,Y), wall(X,Y+1), not wall(X,Y), not wall(X+1,Y+1).
\end{verbatim}
\noindent
Note that the guess whether a cell is a wall or empty in the program
$\progRef{ex-ref-ex:step-.gr}\cup\progRef{ex-ref-ex:mazeNoAS-b.gr}\cup\progRef{ex-ref-ex:mazeUnderstanding-.gr}$
is realised by guessing for each non-border cell whether it is a wall or not
and deriving that a cell is empty in case we do not know that it is a wall.
Moreover, observe that facts of predicate \verb|empty/2| may be part of a valid encoding of a maze generation problem instance, \iec
they are a potential input of the program.
As a consequence, it seems plausible that the encoding could guess the existence of a wall for a cell that is already defined to be empty
by a respective fact in the program input.
In particular, there is no constraint that explicitly forbids that a single cell can be empty and contain a wall.
The encoding would be incorrect if it would allow for answer sets with cells that are empty and a wall according to
the maze generation problem specification.
However, it turns out that the answer sets of the program are exactly the intended ones.
Let us find out why by means of stepping.

\begin{figure}[t]
{
\centering
\includegraphics[width=0.3\textwidth]{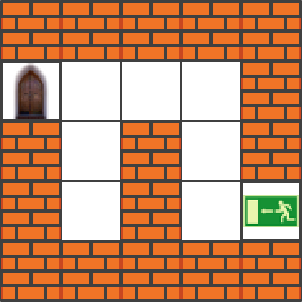}
\caption{A valid maze---but not a solution for the instance depicted in Fig.~\ref{fig:mazeExample} as that requires $(3,4)$ to be an empty cell.}
\label{fig:mazeSolutionWrongInstance}

}
\end{figure}

\begin{figure}[t]
{\centering
\includegraphics[width=0.3\textwidth]{mazeInstance.png}
\hfill
\includegraphics[width=0.3\textwidth]{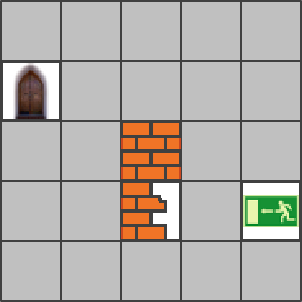}
\hfill
\includegraphics[width=0.3\textwidth]{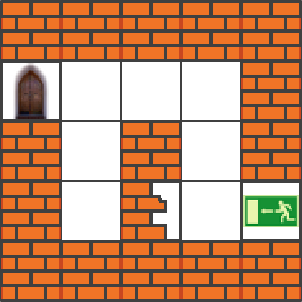}
\caption{The stepping session described in Example~\ref{ex:mazeUnderstanding}: Starting from the maze generation instance we step towards an interpretation encoding the wrong solution of Fig.~\ref{fig:mazeSolutionWrongInstance}.
After stepping through the guessing rule the resulting interpretation contains atoms \texttt{empty(3,4)} and \texttt{wall(3,4)} stating that cell $(3,4)$ is both a wall and empty.}
\label{fig:mazeSteppingTowardsWrongInstance}

}
\end{figure}

Reconsider the problem instance depicted in Fig.~\ref{fig:mazeExample} that is encoded in the program \progRef{exMazeInstance.gr}.
It requires that cell $(3,4)$ is empty.
If it did not, the maze shown in Fig.~\ref{fig:mazeSolutionWrongInstance} that contains a wall at cell $(3,4)$ would be a valid solution.
We start a stepping session for program $\progRef{exMazeInstance.gr}\cup\progRef{ex-ref-ex:step-.gr}\cup\progRef{ex-ref-ex:mazeNoAS-b.gr}\cup\progRef{ex-ref-ex:mazeUnderstanding-.gr}$, whose code is summarised in Fig.~\ref{fig:encodingReminder},
and step towards an interpretation encoding the maze of Fig.~\ref{fig:mazeSolutionWrongInstance}
to see what is happening if we consider $(3,4)$ to be a wall despite the presence of fact \verb|empty(3,4)|.
We can reuse the computation $C_4$ obtained in Example~\ref{ex:jump} whose final state $S_4$ considers already 
the facts describing the input instance and the rules needed for deriving \verb|border/2| atoms.
As in Example~\ref{ex:maze_AS}, we continue with a step for considering the ground instance of the rule
\begin{verbatim}
{wall(X,Y) : col(X), row(Y), not border(X,Y)}.
\end{verbatim}
\noindent
that guesses whether non-border cells are walls.
This time, instead of choosing \verb|wall(3,2)| to be true, we only add \verb|wall(3,4)| to the atoms considered true.
Then, for the resulting state $S'_5$, both \verb|empty(3,4)| and \verb|wall(3,4)| are contained in $\stateI{S'_5}$.
A visualisation of $\stateI{S'_5}$ if given in the centre of Fig.~\ref{fig:mazeSteppingTowardsWrongInstance}.
In order to derive the remaining atoms of predicates \verb|empty/2| and \verb|wall/2| we then jump through the rules
\begin{verbatim}
wall(X,Y) :- border(X,Y), not entrance(X,Y), not exit(X,Y).
empty(X,Y) :- col(X), row(Y), not wall(X,Y).
\end{verbatim}
\noindent
to obtain state $S'_6$, where $\stateI{S'_6}$ is illustrated in the right subfigure of Fig.~\ref{fig:mazeSteppingTowardsWrongInstance}.

Now, the user sees that constraint
\begin{verbatim}
:- empty(X,Y), not reach(X,Y).
\end{verbatim}
\noindent
has active instances.
This comes as no surprise as the rules defining reachability between empty cells have not been considered yet.
We decide to do so now and initiate a jump through the rules
\begin{verbatim}
adjacent(X,Y,X,Y+1) :- col(X), row(Y), row(Y+1).
adjacent(X,Y,X,Y-1) :- col(X), row(Y), row(Y-1).
adjacent(X,Y,X+1,Y) :- col(X), row(Y), col(X+1).
adjacent(X,Y,X-1,Y) :- col(X), row(Y), col(X-1).
reach(X,Y)   :- entrance(X,Y), not wall(X,Y).
reach(XX,YY) :- adjacent(X,Y,XX,YY), reach(X,Y), not wall(XX,YY).
\end{verbatim}
\noindent
We obtain the new state $S'_7$ and observe that 
under interpretation $\stateI{S'_7}$ the constraint still has an active instance, namely
\begin{verbatim}
:- empty(3,4), not reach(3,4).
\end{verbatim}
Obviously, the atom \verb|reach(3,4)| has not been derived in the computation.
When inspecting the rules defining \verb|reach/2| it becomes clear why the answer sets of the encoding are 
correct: 
the atom \verb|reach(X,Y)| is only derived for cells that do not contain walls.
Consequently, whenever there is an empty cell which was guessed to contain a wall
it will be considered as not reachable from the entrance.
As every empty cell has to be reachable, a respective answer-set candidate will be eliminated by an instance of the constraint
\begin{verbatim}
:- empty(X,Y), not reach(X,Y).
\end{verbatim}
Although the encoding of the maze generation problem is correct one could consider it to be not very well designed.
Conceptually, the purpose of the constraint above is forbidding empty cells to be unreachable from the entrance
and not forbidding them to be walls.
Moreover, if one would replace the rules
\begin{verbatim}
reach(X,Y)   :- entrance(X,Y), not wall(X,Y).
reach(XX,YY) :- adjacent(X,Y,XX,YY), reach(X,Y), not wall(XX,YY).
\end{verbatim}
by
\begin{verbatim}
reach(X,Y)   :- entrance(X,Y), empty(X,Y).
reach(XX,YY) :- adjacent(X,Y,XX,YY), reach(X,Y), empty(XX,YY).
\end{verbatim}
\noindent
which seem to be equivalent in the terms of the problem specification, the program would not work.
A more natural encoding would be to explicitly forbid empty cells to contain walls
either by an explicit constraint or a modified guess where non-border cell is guessed to be either empty or contain a wall but not both.
\eex

\section{Related Work}\label{sec:related}
First, we describe existing approaches for debugging answer-set programs
and how they are related to the method proposed in this paper.

The first work devoted to debugging of answer-set programs
is a paper
by \citeauthorandyear{BD05}
in which they provide general considerations on the subject,
such as the discussion of error classes in the context of ASP
or implications of declarativity on debugging mentioned in the previous section.
They also 
formulated important debugging questions in ASP, namely,
why is a set of atoms subset of a specific answer set and
why is a set of atoms not subset of any answer set.
The authors
provided pseudocode for two imperative ad-hoc algorithms for 
answering these questions for propositional normal answer-set programs.
The algorithm addressing the first question returns answers in
terms of active rules that derive atoms from the given set.
The algorithm for explaining
why a set of atoms is not subset of any answer set
identifies different sorts of answers such as
atoms with no deriving rules, inactive deriving rules,
or supersets of the given set in which adding further literals would lead to some inconsistency.

The goal of the work by \citeauthorandyear{PSE09} 
is to explain the truth values of literals with respect to a given actual answer set of a program.
Explanations are provided in terms of \memph{justifications} which are labelled graphs
whose nodes are truth assignments of possibly default-negated ground atoms.
The edges represent positive and negative support relations between these truth assignments such that every path ends in an assignment which is either assumed or known to hold.
The authors have also introduced justifications for partial answer sets that emerge during the solving process (online justifications), being represented by three-valued interpretations.
\citeauthorandyear{PSE09} use sequences of three-valued interpretations (called computations)
in which monotonously more atoms are considered true, respectively, false. 
The information carried in these interpretations corresponds to that of the second and third component of a state in 
a computation in our framework.
The purpose of using computations in the two approaches differs however.
While computations in stepping are used for structuring debugging process in a natural way, where the choices how to proceed remains
with the user,
the computations of \citeauthor{PSE09}
are abstractions of the solving procedure.
Their goal is to allow a solver that is compatible with their computation model to compute justifications 
for its intermediate results.
Thus, similar to their offline versions, online justifications are non-interactive, \iec they are computed automatically and used for
post-mortem debugging.
As our computation model is compatible with that for online justifications, it seems
very promising to combine the two approaches in practice.
While debugging information in stepping focuses on violation of rules and unfounded sets,
our method leaves the reasons for an atom being true or false as implicit consequences of a user's decision.
Here, online justifications could keep track of the reasons for truth values at each state of a stepping session
and presented to the user during debugging on demand.

\citeauthorandyear{S06} aimed at finding explanations why a program has no answer sets. 
His approach is based on finding minimal sets of constraints such that their removal yields consistency.
Hereby, it is assumed that a program does not involve circular dependencies
between literals through an odd number of negations which might also cause inconsistency.
The author considers only a basic ASP language and hence
does not take further sources of inconsistency into account, caused by program constructs of richer ASP languages, such as cardinality constraints.

Another early approach~\citep{BGPSTW07,P07} is based on
program rewritings using some additional control
atoms, called \emph{tags}, that allow, \egc for switching individual rules on or off and for analysing the resulting answer sets.
Debugging requests can be posed by adding further rules that can employ tags as well.
That is, ASP is used itself for debugging answer-set programs.
The translations needed were implemented in the
command-line tool \spock~\citep{BGPSTW07b,GPSTW09}
which also incorporates the translations of
another approach in which also ASP is used for debugging purposes~\citep{GPST08,P07}.
The technique is based on ASP meta-programming,
\iec a program over a meta-language is used to manipulate a program over an object language (in this case, both the meta-language and the object language are instances of ASP).
It addresses the question why some interpretation is not an answer set of the given program.
Answers are given in terms of unsatisfied rules and unfounded loops.
The approach has later been extended from propositional to
disjunctive logic programs with  constraints,
integer arithmetic, comparison predicates, and strong negation~\citep{OPT10}
and also to programs with cardinality constraints~\citep{PFSF13}.
It has been implemented in the \ouroboros plugin of \sealion~\citep{FPF13}.
Moreover, \citeauthorandyear{S15} developed a method on top of the meta-programming approaches \citep{GPST08,OPT10}
that poses questions to the user in order to find a desired problem diagnosis while keeping the amount of required interaction low.

In a related approach,
\citeauthorandyear{DGMRS15} use control atoms quite similar to that of the tagging approach~\citep{BGPSTW07,P07}
to identify sets of rules that lead to inconsistency of a program under the requirement that a given set of atoms is true in some intended answer set.
An implementation is provided that profits from a tight integration with the ASP solver \wasp~\citep{AlvianoDLR15}. In order to reduce the possible outcomes, the debugger asks the user about the intended truth of further atoms in an interactive session.

In another paper, \citeauthorandyear{LiVPSB15} use inductive logic programming to repair answer-set programs. The idea is that the user provides examples of desired and undesired properties of answer sets such that the debugger can semi-automatically revise a faulty program. The method requires a difference metric between logic programs in order to restrict repairs to programs that have the desired properties that minimally differ from the original program. The authors propose such a measure in terms of number of operations required for revision. These operations are rule removal and creation as well as addition or removal of individual body literals.

\citeauthorandyear{CGS08} developed a declarative debugging approach for datalog
using a classification of error explanations similar to that of the aforementioned meta-programming technique~\citep{GPST08,OPT10}.
Their approach is tailored towards query answering  and the language is restricted to stratified datalog. 
However, the authors provide an implementation that is based on computing
a graph that reflects the execution of a query.

\citeauthorandyear{WVD09} show how a calculus can be used for
debugging  first-order theories with inductive definitions~\citep{D00,DT08} in the context of model expansion problems, \iec problems of finding models of a given theory that expand some given interpretation. 
The idea is to trace the proof of inconsistency of such an unsatisfiable model expansion problem.
The authors provide a system that allows for interactively exploring the proof tree.

Besides the mentioned approaches which rely on the semantical behaviour of programs, \citeauthorandyear{MMT07} use a translation from logic-program rules to natural language in order to detect program errors more easily.
This seems to be a potentially useful feature for an IDE as well, especially for non-expert ASP programmers.

We can categorise the different methods by their setting of their debugging tasks.
Here, one aspect is whether a technique works with factual or desired answer sets.
Approaches that focus on actual answer sets of the program to be debugged
include the algorithm by \citeauthorandyear{BD05} that
aims at explaining the presence of atoms in an answer set.
Also, justifications~\citep{PSE09} are targeted towards explanations in a given actual answer set,
with the difference that they focus on a single atom but can not only explain their presence
but also their absence.
The approach by \citeauthorandyear{CGS08}
can also be seen to target a single actual answer set.
Due to their focus on actual answer sets of the debugged program, these methods
cannot be applied on (erroneous) programs without any answer set.
The previous meta-programming based debugging technique~\citep{GPST08,P07} and follow-up works~\citep{OPT10,PFSF13}
deal with a single intended but non-actual answer set of the debugged program.
In the approach of \citeauthorandyear{WVD09}, the user can specify a class of intended semantic structures which are not preferred models of the theory at hand (corresponding to actual answer sets of the program to be debugged in ASP terminology).
\citeauthor{S06}'s diagnosis technique~\citep{S06} is limited to the setting when a program has no answer set at all. The same holds for the work of \citeauthorandyear{DGMRS15}, however the authors demonstrate how other debugging problems can be reduced to that of inconsistency.
The method requires an intended answer set but offers the means to generate that in an interactive way, building on the technique by \citeauthorandyear{S15}.
Stepping does not require actual or intended answer sets as a prerequisite,
as the user can explore the behaviour of his or her program under different interpretations
that may or may not be extended to answer sets by choosing different rules instances.
In the interactive setting summarised in Fig.~\ref{fig:steppingCycle}, where one can
retract a computation to a previous state and continue stepping from there
that is also implemented in \sealion,
a stepping session can thus be seen as an inspection across arbitrary interpretations rather than
an inquiry about a concrete set of actual or non-existent answer sets.
Nevertheless, if one has a concrete interpretation in mind, the user is free to focus on that.
The ability to explore rule applications
for partial interpretations that cannot become answer sets
amounts to a form of hypothetical reasoning.
A related form of this type of debugging is also available in one feature of the tagging approach~\citep{BGPSTW07}
that aims at extrapolating non-existent answer sets by switching off rules and guessing further atoms.
Here, the stepping technique can be considered more focused, as the 
interpretation under investigation is determined by the choices of the user in stepping
but is essentially arbitrary in the tagging approach if the user does not employ explicit restrictions.

Next, we compare the ASP languages supported by different approaches.
First, the language of theories with inductive definitions used in one of the debugging approaches~\citep{WVD09}
differs from the remaining approaches that are based on logic-programming rules.
Many of these works deal only with the basic ASP setting of
debugging ground answer-set programs,
supporting only normal rules~\citep{BD05,PSE09},
disjunctive rules~\citep{GPST08},
or simple choice rules~\citep{S06}.
The work on tagging-based debugging~\citep{BGPSTW07} sketches how to 
apply the approach to programs with variables by means of function symbols.
The approach by \citeauthorandyear{CGS08}
deals with non-ground normal programs which have to be stratified.
Explicit support for variables 
is also given in an extension~\citep{OPT10} of the meta-programming approach
for disjunctive programs.
It was later extended to allow for weight constraints~\citep{PFSF13}
by compiling them away to normal rules.
A commonality of the previous approaches is that they target ASP
languages that can be considered idealised proper subsets of current solver languages.
In this respect, stepping is the first debugging approach that
overcomes these limitations as the use of \cprograms and abstract grounding (cf.~\citealp{Puehrer14})
make the framework generic enough to be applied to ASP solver languages.
While this does not mean that other approaches cannot be adapted to fit a solver language,
it is no always immediately clear how.
For our approach, instantiating our abstractions to the language constructs and the grounding method of a solver
is sufficient to have a ready-to-use debugging method.

Most existing debugging approaches for ASP can be seen as declarative
in the sense that a user can pose a debugging query,
and receives answers in terms of different declarative definitions of the semantics
of answer-set programs, \egc in terms of active or inactive rules with respect to some interpretation.
In particular, the approaches do not take the execution strategy of solvers into account.
This also holds for our approach,
however stepping and online justifications~\citep{PSE09} are exceptional
as both involve a generic notion of computation
which adds a procedural flavour to debugging.
Nonetheless, the computation model we use can be seen as a declarative
characterisation of the answer-set semantics itself as it does not apply a fix order in which to apply rules
to build up an answer set.

Besides stepping, also 
the approaches by \citeauthorandyear{WVD09} and \citeauthorandyear{DGMRS15} as well as \citeauthorandyear{S15}
can be considered interactive.
While in the approach of \citeauthor{WVD09} a fixed proof is explored interactively,
the interaction in our method has influence on the direction of the computation.
The other works~\citep{DGMRS15,S15} use interaction for filtering
the resulting debugging information.
Also in further works~\citep{BGPSTW07,GPST08} 
which do not explicitely cover interleaved communication between
user and system, user information can be used for filtering.
The approaches mentioned in this paragraph realise declarative debugging
in the sense of \citeauthorandyear{S82}, where the user serves as an oracle for guiding
the search for errors.

It is worth highlighting that stepping can be seen as orthogonal to the basic ideas of all the other approaches we discussed.
That is, it is reasonable to have a development kit that supports stepping and other 
debugging methods simultaneously.

While debugging is the main focus of this paper, we also consider 
the computation framework for disjunctive abstract constraint programs introduced in Section~\ref{sec:computation}
an interesting theoretical contribution by itself.
Here, an important related work is that of
\citeauthorandyear{LPST10}, who also use a notion of computation to characterise
a semantics for normal \cprograms.
These computations
are sequences of evolving interpretations.
Unlike the three-valued ones used for online justifications \citep{PSE09},
these carry only information about atoms considered true.
Thus, conceptionally, they correspond to sequences 
$\stateI{S_0},\stateI{S_1},\dots$ where
$S_0,S_1,\dots$ is a computation in our sense.
The authors formulate principles for characterising different variants of computations.
We will highlight differences and commonalities between the approaches along the lines
of some of these properties.
One main structural difference between their and our notion of computation is the granularity of steps:
In the approach by \citeauthor{LPST10} it might be the case that multiple rules must be considered
at once, as required by their \memph{revision property} (R'),
while in our case computation proceeds rule instance by rule instance.
The purpose of property (R') is to assure 
that every successive interpretation
must be supported by the rules active with respect to the previous interpretation.
But it also requires that every active rule in the overall program is satisfied
after each step, whereas we allow rule instances that
were not considered yet in the computation to be unsatisfied.
For the purpose of debugging, rule-based computation granularity seems favourable
as rules are our primary source code artifacts.
Moreover, ignoring parts of the program that were not considered yet in a computation
is essential in the stepping method, as this breaks down the amount of information
that has to be considered by the user at once and allows for getting stuck and
thereby detect discrepancies between his or her understanding of the program and its
actual semantics.
Our computations (when translated as above) meet the \memph{persistence principle} (P')
of \citeauthor{LPST10} that ensures that a successor's interpretation is a superset
of the current one.
Their \memph{convergence principle} (C'),
requiring that a computation stabilises to a supported model,
is not met by our computations, as we do not enforce support in general.
However, when a computation has succeeded (\cf\  Definition~\ref{def:computationsemantics2}), it meets this property.
A further difference is that \citeauthor{LPST10} do not allow for non-stable computations
as required by the founded persistence of reasons principle (FPr).
This explains why the semantics they characterise treats non-convex atoms
not in the FLP-way.
Besides that, the use of non-stable computations allow us to handle disjunction.
Interestingly, \citeauthor{LPST10} mention the support for disjunction in computations as an open
challenging problem and suspect the necessity of a global minimality requirement
on computations for this purpose.
Our framework demonstrates that we can do without such a condition:
As shown in Theorem~\ref{th:extSuppLocally}, unfounded sets
in our semantics can be computed incrementally ``on-the-fly'' by considering only
the rule instance added in a step as potential new external support.
Finally, the \memph{principle of persistence of reasons} (Pr')
suggests that the ``reason'' for the truth value of an atom
must not change in the course of a computation.
\citeauthor{LPST10} identify such reasons by sets of rules that keep providing support in an ongoing computation.
We have a similar principle of persistence of reasons that is however stricter as it is operates on the atom level rather than the rule level:
Once a rule instance is considered in a computation in our sense, the truth value of the atoms in the rule's domain
is frozen, \iec it cannot be changed or forgotten in subsequent steps.
Persistence of reasons is also reflected in our definition of answer sets:
The requirement $\proj{\inter'}{\domain{A}}=\proj{\inter}{\domain{A}}$ in Definition~\ref{def:answersets}
that the stability of interpretation $I$ is only spoiled by $I'$ if the reason for
$\inter'\models A$  is the same satisfier of \catom $A$ as for $\inter\models A$.

As argued above and in the introduction, our notion of computation is better suited for stepping than that of \citeauthor{LPST10},
yet we see potential for using the latter orthogonal to our method for debugging (for the class of programs for which the different semantics coincide).
While our jumping technique allows to consider several rules, selected by the user, at once, a debugging system could provide proposals for jumping,
where each proposal corresponds to a set of rules that result in a next step in the sense of \citeauthor{LPST10}
Then, the system could present the atom assignments for each proposal such that the user has an alternative to choose a jump based on truth of atoms rather than rules.
Moreover, this can be the basis for automated progression in a stepping session until a certain atom is assigned, analogous to watchpoints in imperative debugging.
We believe that these ideas for (semi-)automated jumping deserve further investigation.

Another branch of research, that is related to our notion of computation, focuses on transition systems for analysing answer-set computation~\cite{L11,LierlerT16,BrocheninLM14}.
These works build on the ideas of a transition system for the DPLL procedure for SAT solving~\cite{NieuwenhuisOT06}.
Transition systems are graphs whose nodes represent the state of a computation whereas the edges represent possible state transitions during solving.
Typically, a set of transition rules, depending on the answer-set program, determines the possible transitions.
In ASP transition systems, nodes are represented by (ordered) sets of literals with annotations whether a literal is a decision literal. Moreover, there is a distinguished state 
$\mathit{FailState}$
for representing when a branch of computation cannot succeed in producing an answer set.
Different sets of transition rules have been proposed that correspond to different models of computations. Typical transition rules include a unit propagation rule that derives one further literal based on a rule of the program for which all other literals are defined in the previous state, a decision rule that adds one further literal (annotated as decision literal), and a transition rule for backtracking that retracts a decision literal from an inconsistent state and replace it by its negation.
Existing transition systems for ASP are intended to reflect ASP solving algorithms, including failed branches of search space traversal. For instance, transition systems have been defined with transition rules for backjumping and learning as used in modern solvers~\cite{L11}.
In contrast, our framework generates ideal (possibly failed) computations without backtracking.
Another main difference is that all proposed transition systems have a transition rule for
arbitrary assignment of decision literals whereas in our framework truth assignments are restricted to the domain of the \crule added in the current step.
Regarding supported language constructs, to the best of our knowledge, existing transition systems for ASP focus on elementary atoms, \iec they do not cover aggregates. However, \citeauthorandyear{LierlerT16} also proposed transition systems for multi-logic systems including ASP.
There has been work on transition systems for disjunctive programs~\cite{BrocheninLM14}. These are based on integrating two sets of transition rules, one for guessing and one for checking of answer set candidates.
Similarly, as in the work by \citeauthorandyear{LPST10}, states in transition systems do not keep track of ASP rules as our states do.
Note that our computation framework can be turned into to a transition system for disjunctive \cprograms 
with only two transition rules, one for propagation that is derived from the successor relation (\cf Definition~\ref{def:successor}) 
and another for defining the transition of unstable final states or states with remaining active rules but no successor to $\mathit{FailState}$.

\section{Conclusion}\label{sec:conclusion}
In this paper, we introduced the stepping technique for ASP
that can be used for debugging and analysis of answer-set programs.
Like stepping in imperative programming, where the effects of 
consecutive statements are watched by the user,
our stepping technique allows for monitoring the effect of
rules added in a step-by-step session.
In contrast to the imperative setting, stepping in our sense
is interactive as a user decides in which direction to branch,
by choosing which rule to consider next and which truth values its atoms should be assigned.
On the one hand, this breaks a general problem of debugging in ASP, namely how to find the cause for an error,
into small pieces.
On the other hand, user interaction allows for focusing on interesting parts of the debugging search space from the beginning.
This is in contrast to the imperative setting, where the order in which statements are considered in a debugging session is fixed.
Nevertheless, also in our setting, the choice of the next rule is not entirely arbitrary,
as we require the rule body to be active first.
Debuggers for procedural languages often tackle the problem
that many statements need to be considered before coming to an interesting step
by ignoring several steps until pre-defined breakpoints are reached.
We developed an analogous technique in our approach that we refer to as jumping which allows to consider multiple rules at once.
Besides developing the technical framework for stepping, we also discussed 
the implementation of stepping in the
\sealion system
and methodological aspects, thereby giving guidelines for the usage of the technique, and for
setting the latter in the big picture of ASP development.

While unstable computations are often not needed, they offer great opportunities
for further work.
For one, the use of unfounded sets for distinguishing states in unstable computations is a natural 
first choice for expressing the lack of stability.
Arguably, when a user arrives in a state with a non-empty unfounded set, he or she 
only knows that some external support has to be found for this set but there is no information
which atoms of the unfounded sets are the crucial ones.
It might be worthwhile to explore alternative representations for unstability
such as elementary loops~\citep{GLL11} that possibly provide more pinpoint information.
This would require lifting a respective notion to the full language of \cprograms first.

Another issue regarding unstable computations that would deserve further attention
is that in the current approach jumps can only result in stable states.
Thus, unstable states in a computation can only be reached by individual steps at present.
Here, it would be interesting to study methods and properties for computations that allow
for jumping to states that are not stable.

We next discuss functionality that could be helpful
for stepping which are not yet implemented in \sealion.
One such feature is semi-automatic stepping, \iec the user can push a button and
then the system searches for potential steps for which no further user interaction is required
and applies them automatically until an answer set is reached, the computation is stuck,
or user interaction is required.
It would also be convenient to automatically check whether the computation of a debugging session
is still a computation for the debugged program after a program update.
In this respect, when the computation for the old version became incompatible, 
a feature would be advantageous that builds up a computation for the new version that
resembles the old one as much as possible.
Unlike semi-automatic stepping and compatibility checks for computations which could be implemented without further studies, the latter point still requires theoretical research.
Further convenient features would be functionality that 
highlights the truth values of atoms that cause a rule not to be active for a given substitution
and methods for predicting whether a rule can become active in the future, \iec in some continuation of the computation.

\section*{Acknowledgements}\label{sec-ackn}
We thank the reviewers for their useful comments.
This work was partially supported by the Austrian Science Fund (FWF) under project P21698,
the German Research Foundation (DFG) under grants BR-1817/7-1 and BR 1817/7-2,
and the European Commission under project IST-2009-231875 (OntoRule).

\ifCORR
\appendix
\section{Guidelines for Stepping}\label{sec:guidelines}

In what follows, we give advice on how users can exploit stepping for analysing and debugging their code.
Fig.~\ref{fig:steppingGuide} synthesises practical guidelines for stepping
from the methodological aspects of stepping described in Section~\ref{sec:methodology}.
It can be seen as a user-oriented view on the stepping technique.
Depending on the goals and the knowledge of the user,
this guide gives concise yet high-level suggestions on how to proceed in a stepping session.
The upper area of the figure is concerned with clarifying the best strategy
for a stepping session and for choosing the computation to start from.
The lower area, on the other hand, guides the user through the stepping process.

The diagram differentiates between four tasks a user may want to perform.
\bi
\item[(i)] Debugging a program lacking a particular answer set:
       we suggest to step and jump through rules
       that one thinks build up this answer set.
       Eventually, the computation will get stuck when adding a rule that prevents the answer set.
\item[(ii)] Debugging a program that lacks any answer set:
       if an intended answer set is known, we advise using the strategy of Item (i).
       Otherwise, the user should choose rules and truth values during stepping that
       he or she thinks should be consistent, \iec lead to a successful computation.
       Also here, the computation is guaranteed to fail and get stuck, indicating
       a reason for the inconsistency.
\item[(iii)] Debugging a program with an unintended answer set $I$:
       In case that $I$ is similar to an intended but missing answer set $I'$, thus
       if $I$ is intuitively a wrong version of $I'$, we recommend stepping towards $I'$, following
       the strategy of Item (i). 
       Otherwise, the user can step towards $I$. Unlike in the previous cases, the computation is guaranteed
       to eventually succeed. Here, stepping acts as a disciplined way to inspect how the atoms of $I$ can 
       be derived and why no rule prevents $I$ from being an answer set. If $I$ is intended to be a model
       but not stable, then the stepping process will reveal which rules provide external support for sets
       of atoms that are supposed to be unfounded.
       
\item[(iv)] Analysing a program: In case that the user is interested in the behaviour of the program under a particular
            interpretation, it is reasonable to step towards this interpretation. 
            Otherwise, rules and truth assignments should be chosen that drive the computation
            towards states that the user is interested in.
\ei
\noindent
The procedures suggested above and in Fig.~\ref{fig:steppingGuide} are meant as rough guidelines
for the inexperienced user.
Presumably, knowledge about the own source code and some practice in stepping gives the user a good intuition
on how to find bugs efficiently.

\afterpage{\begin{figure}[p]
{\centering
\includegraphics[width=\textwidth]{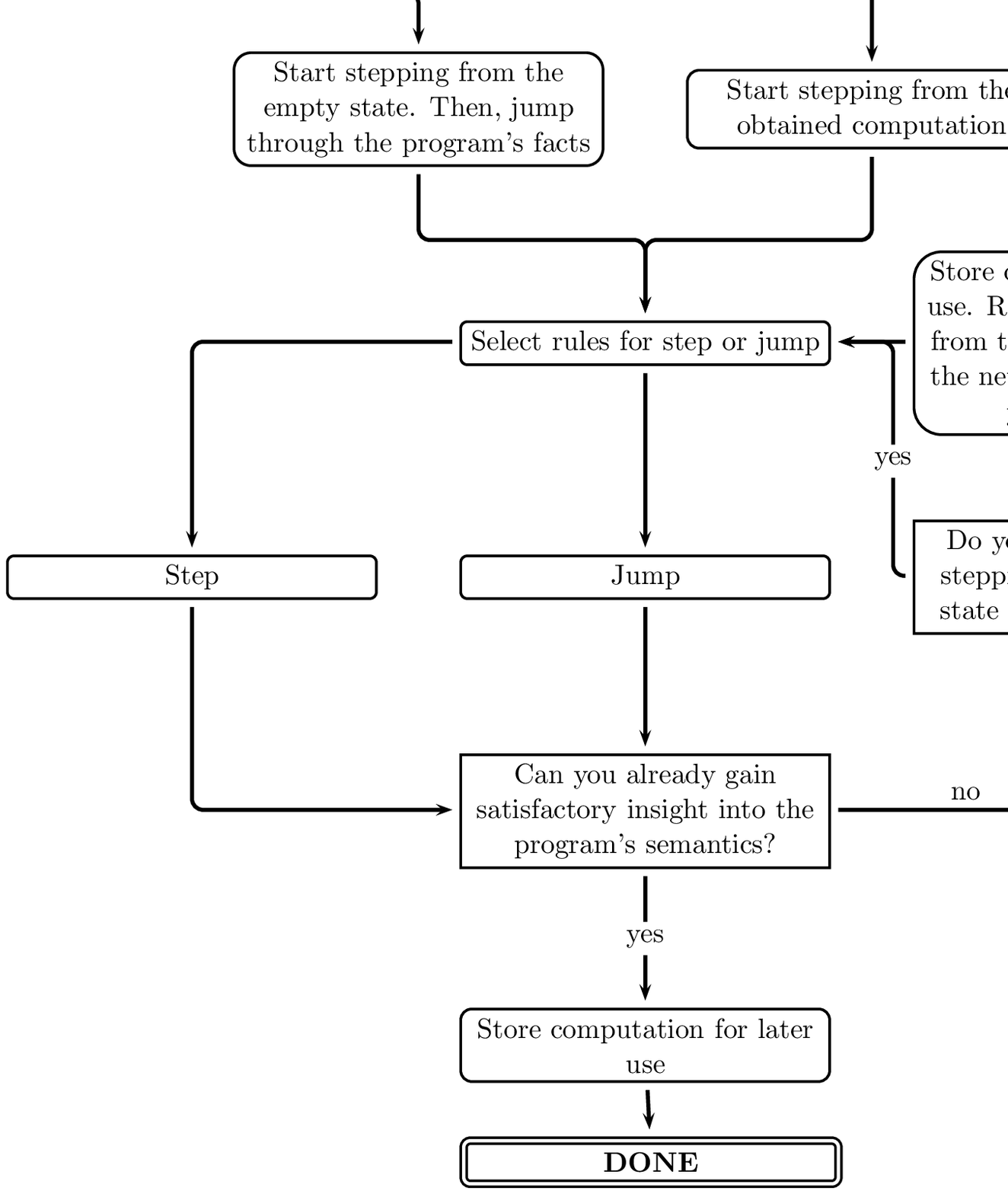}
\caption{Stepping guide}
\label{fig:steppingGuide}

}
\end{figure}
\clearpage
}

It is natural to ask how big a program can get such that it is still suitable for stepping.
Due to the vague nature of the question, answers cannot be clearly established.
From a complexity theoretic point of view, 
the problems that need to be solved in a stepping support environment
for and after performing a step or a jump, \egc computing a new state from a jump, determining rules with active instances, or checking whether a computation has failed, 
are not harder than computing an answer set of the program under development.
Under this observation, our technique is certainly an appropriate approach for debugging ASP.
In some applications, however, solving times of multiple minutes or even hours are acceptable.
Certainly, having waiting times of these lengths for individual debugging steps is undesirable.
On the positive side, often, following a few guidelines during the development of an answer-set program can significantly
reduce the likelihood of introducing bugs, the amount of information the user has to deal with,
and also the computational resources required for stepping.
Among these measures are best practices for ASP development that have been discussed in a paper by \citeauthorandyear{BCD09}.
For working with the stepping method in particular, we give the following recommendations.

\paragraph{Use scalable encodings and start with small examples.}
Using small problem instances, also the resulting grounding as well as answer sets are typically small.
This limits the amount of information to be considered during debugging.
Chances that bugs are detected early, using small programs is suggested by an evaluation of the \emph{small-scope hypothesis} for ASP~\citep{OPPST12}.

\paragraph{Visualise answer sets and stepping states.}
Tools like \kara~\citep{KOPT13} (that is implemented in \sealion), \aspviz~\citep{CDBP08}, \idpdraw~\citep{W09}, or \lonsdaleite~\citep{S11} allow for
visualising interpretations.
With their help, one can quickly spot when an answer set differs from what is expected and they allow to monitor the evolvement of the interpretation that is build up during stepping.
The illustrations of the maze generation problem in this section were created using \kara.
For use with stepping, we advise to specify visualisations also for interpretations that are not supposed to be answer sets.
For example, in Fig.~\ref{fig:mazeSteppingTowardsWrongInstance}, we have visualisations for cells that are not assigned to be empty or a wall and for cells that are assigned to be a wall and empty, despite
in an expected answer set, every cell has to be either a wall or empty.

\paragraph{Test often.}
Frequent tests allow the user to trust in large parts of the program, hence
these parts can be jumped over in a stepping session.

\section{Remaining Proofs}\label{sec:proofs}

\begin{theoremnr}{\ref{th:extSuppLocally}}
Let $S$ be a state 
and $S'$ a successor of $S$,
where $\Delta=\stateI{S'}\setminus\stateI{S}$. 
Moreover, let $X'$ be a set of literals with $\emptyset\subset X'\subseteq \stateI{S'}$.
Then, the following statements are equivalent:
\bi
\item[{\rm (}i{\rm )}] 
$X'$ is unfounded in $\stateprog{S'}$ with respect to $\stateI{S'}$.
\item[{\rm (}ii{\rm )}] $X'=\Delta'\cup X$, where $\Delta'\subseteq \Delta$, $X\in\stateunfounded{S}$, and
$\newRule{S}{S'}$ is not an external support for $X'$ with respect to $\stateI{S'}$.
\ei
\end{theoremnr}
\bp
((i)$\Rightarrow$(ii))
It is obvious that $\newRule{S}{S'}$ is not an external support for $X'$ with respect to $\stateI{S'}$
as otherwise $X'$ cannot be unfounded in $\stateprog{S'}$ with respect to $\stateI{S'}$.
It remains to be shown that $X'=\Delta'\cup X$ for some $\Delta'\subseteq \Delta$ and some $X\in\stateunfounded{S}$.
Towards a contradiction, assume $X'\neq \Delta'' \cup X''$ for all $X''\in\stateunfounded{S}$ and all $\Delta''\subseteq\Delta$.
We define $X=X'\cap \stateI{S}$.

Consider the case that $X\in\stateunfounded{S}$.
As $X'\setminus \stateI{S}\subseteq\Delta$,
and $X'= (X'\setminus \stateI{S}) \cup X$, we have a contradiction
to our assumption.
Therefore, it holds that $X\not\in\stateunfounded{S}$.
Hence, as $X\subseteq \stateI{S}$,
by definition of a state, $X$ is not unfounded in $\stateprog{S}$ with respect to $\stateI{S}$.
Therefore, there is some external support $r\in\stateprog{S}$ for $X$ with respect to $\stateI{S}$.

In the following, we show that $r$ is also an external support for $X'$ with respect to $\stateI{S'}$.
Since $S'$ is a successor of $S$ and $S$ is a state,
we get that $\stateI{S}$ and $\stateI{S'}$ coincide on $\domain{r}$.
Consequently, from $\stateI{S}\models\body{r}$  we get that also $\stateI{S'}\models\body{r}$.
Moreover, because of $\stateI{S}\setminus X \models\body{r}$ it is also true that
$\stateI{S'}\setminus X' \models\body{r}$.
Furthermore, we know that
there is some $A\in\head{r}$ with
$\proj{X}{\domain{A}}\neq\emptyset$ and $\proj{\stateI{S}}{\domain{A}}\subseteq C$, for some $C\in\satisfiers{A}$.
As $\proj{X}{\domain{A}}=\proj{X'}{\domain{A}}$ and $\proj{\stateI{S}}{\domain{A}}=\proj{\stateI{S'}}{\domain{A}}$
we also have
$\proj{X'}{\domain{A}}\neq\emptyset$ and $\proj{\stateI{S'}}{\domain{A}}\subseteq C$.
Finally, note that
for all $A\in\head{r}$ with $\stateI{S}\models A$, we have $\proj{(X\cap \stateI{S})}{\domain{A}}\neq\emptyset$.
Consider some $A\in\head{r}$ such that $\stateI{S'}\models A$.
From the latter we get that $\stateI{S}\models A$ and therefore
$\proj{(X\cap \stateI{S})}{\domain{A}}\neq\emptyset$.
As $X\cap \stateI{S} \subseteq X'\cap \stateI{S'}$, we also have $\proj{(X'\cap \stateI{S'})}{\domain{A}}\neq\emptyset$.
Hence, $r$ fulfils all conditions for being an external support for $X'$ with respect to $\stateI{S'}$,
which is a contradiction to $X'$ being unfounded in $\stateprog{S'}$ with respect to $\stateI{S'}$.

((ii)$\Rightarrow$(i))
Towards a contradiction, assume $X'$ has some external support $r\in\stateprog{S'}$ with respect to $\stateI{S'}$.
From (ii) we know that $r\neq \newRule{S}{S'}$ and 
$X'=\Delta'\cup X$ for some $\Delta'\subseteq \Delta$ and some $X\in\stateunfounded{S}$.
As $r\neq \newRule{S}{S'}$, we have that $\stateI{S}$ and $\stateI{S'}$ coincide on $\domain{r}$.
Therefore,
from $\stateI{S'}\models\body{r}$ and $\stateI{S'}\setminus X' \models\body{r}$,
it follows that
$\stateI{S}\models\body{r}$ and
$\stateI{S}\setminus X'\models\body{r}$.
Note that $X=X'\cap \stateI{S}$ and hence $\stateI{S}\setminus X\models\body{r}$.
We know that there
is some $A\in\head{r}$ with $\proj{X'}{\domain{A}}\neq\emptyset$ and $\proj{\stateI{S'}}{\domain{A}}\subseteq C$, for some $C\in\satisfiers{A}$.
As $\proj{X'}{\domain{A}}=\proj{X}{\domain{A}}$ we have $\proj{X}{\domain{A}}\neq\emptyset$.
Moreover, as 
$\proj{\stateI{S'}}{\domain{A}}=\proj{\stateI{S}}{\domain{A}}$,
it holds that $\proj{\stateI{S}}{\domain{A}}\subseteq C$.
Finally, notice that 
for all $A\in\head{r}$ with $\stateI{S'}\models A$, we have $\proj{(X'\cap \stateI{S'})}{\domain{A}}\neq\emptyset$.
Consider some $A\in\head{r}$ with $\stateI{S}\models A$.
As $\proj{\stateI{S'}}{\domain{A}}=\proj{\stateI{S}}{\domain{A}}$, we also have 
$\stateI{S'}\models A$ and hence $\proj{(X'\cap \stateI{S'})}{\domain{A}}\neq\emptyset$.
As $\domain{A}\cap\Delta=\emptyset$, we have 
$\proj{(X'\cap \stateI{S'})}{\domain{A}}=\proj{(X\cap \stateI{S})}{\domain{A}}$.
Consequently, it holds that $\proj{(X\cap \stateI{S})}{\domain{A}}\neq\emptyset$.
We showed that $r$ is an external support of $X$ in $\stateprog{S}$ with respect to $\stateI{S}$.
Therefore, we have a contradiction to $X\in\stateunfounded{S}$ because $S$ is a state.
\ep

\begin{theoremnr}{\ref{th:complete}}
Let $S_0$ be a state, $\progC$ a \cprogram with $\stateprog{S_0}\subseteq\progC$,
and $I$ an answer set of $\progC$ with $\stateI{S_0}\subseteq I$ and $I\cap\statenegI{S_0}=\emptyset$.
Then, there is a computation $S_0,\dots,S_n$ that has succeeded for $\progC$ such that
$\stateprog{S_n}=\reductFLP{\progC}{I}$ and $\stateI{S_n}=I$.
\end{theoremnr}
\bp
The proof is by induction on the size of the set $\reductFLP{\progC}{\inter}\setminus\stateprog{S_0}$.
Observe that from
$\stateI{S_0}\subseteq I$, $I\cap\statenegI{S_0}=\emptyset$,
and $\stateI{S_0}\models\body{r}$ and
$\domain{r}\subseteq \stateI{S_0}\cup\statenegI{S_0}$,
for all $r\in\stateprog{S_0}$,
we get that 
$I\models\body{r}$ for all $r\in\stateprog{S_0}$.
Hence, as
$\stateprog{S_0}\subseteq\progC$,
we have
$\stateprog{S_0}\subseteq\reductFLP{\progC}{\inter}$.

Consider the base case that $\abs{\reductFLP{\progC}{\inter}\setminus\stateprog{S_0}}=0$.
From $\stateprog{S_0}\subseteq\reductFLP{\progC}{\inter}$ we get $\stateprog{S_0}=\reductFLP{\progC}{\inter}$.
Consider the sequence $C=\tuple{\stateprog{S_0},\stateI{S_0},\statenegI{S_0},\stateunfounded{S_0}}$.
Towards a contradiction, assume $\stateI{S_0}\neq I$.
As $\stateI{S_0}\subseteq I$ this means $\stateI{S_0}\subset I$.
Hence, there is some $a\in I\setminus \stateI{S_0}$.
As for all $r\in\stateprog{S_0}$ it holds that $\domain{r}\subseteq \stateI{S_0}\cup\statenegI{S_0}$,
and $I\cap\statenegI{S_0}=\emptyset$, we get 
$a\not\in\domain{\stateprog{S_0}}$.
We have a contradiction to $I\in\AS{\stateprog{S_0}}$ by Corollary~\ref{cor:asunfounded}, as
$\{a\}$ is unfounded in $\stateprog{S_0}$ with respect to $I$.
Consequently, $\stateI{S_0}=I$ must hold.
As $\stateI{S_0}$ is an answer set of $\stateprog{S_0}$ and $S_0$ is a state, we have that $\stateunfounded{S_0}=\{\emptyset\}$ by definition of state.
It follows that $C$ meets the criteria of the conjectured computation.

We proceed with the step case.
As induction hypothesis, assume that the claim holds whenever
$\abs{\reductFLP{\progC}{\inter}\setminus\stateprog{S_0}}\leq i$ for an arbitrary but fixed $i\ge 0$.
Consider some state $S_0$ and some $I\in\AS{\stateprog{S_0}}$ for which the conditions
in the premise hold such that
$\abs{\reductFLP{\progC}{\inter}\setminus\stateprog{S_0}}=i+1$.
Towards a contradiction, assume there is no \crule $r\in\reductFLP{\progC}{\inter}\setminus\stateprog{S_0}$ such that $\stateI{S_0}\models\body{r}$.
Note that there is at least one \crule $r'\in \reductFLP{\progC}{\inter}\setminus\stateprog{S_0}$ because $\abs{\reductFLP{\progC}{\inter}\setminus\stateprog{S_0}}=i+1$.
It cannot hold that $I=\stateI{S_0}$
since from $r'\in\reductFLP{\progC}{\stateI{S_0}}$ follows $\stateI{S_0}\models\body{r'}$.
Consequently, we have $\stateI{S_0}\subset I$.
Consider some $r''\in\reductFLP{\progC}{\inter}$ with $\stateI{S_0}\models\body{r''}$.
By our assumption, we get that $r''\in\stateprog{S_0}$.
It follows that $\stateI{S_0}\models r''$, and consequently
there is some \catom $A\in\head{r''}$ with $\stateI{S_0}\models A$.
As $\domain{r''}\subseteq\domain{S_0}$, we have $\domain{A}\subseteq \stateI{S_0}\cup\statenegI{S_0}$.
From that, since $\stateI{S_0}\subset I$ and $I\cap\statenegI{S_0}=\emptyset$,
we get $\proj{I}{\domain{A}}=\proj{\stateI{S_0}}{\domain{A}}$.
We have a contradiction to $I$ being an answer set of $\progC$ by
Definition~\ref{def:answersets}.

So, there must be some \crule $r\in\reductFLP{\progC}{\inter}\setminus\stateprog{S_0}$ such that $\stateI{S_0}\models\body{r}$.
From $r\in\reductFLP{\progC}{\inter}$ we get $I\models\body{r}$ and $I\models r$.
Consider the state structure $S_1=\tuple{\progC_1,I_1,\negpart{I_1},\unfounded_1}$, where
$\progC_1=\stateprog{S_0}\cup\{r\}$,
$I_1=\stateI{S_0}\cup(I\cap\domain{r})$,
$\negpart{I_1}=\statenegI{S_0}\cup(\domain{r}\setminus I)$, and

\[
\begin{array}{l@{}l}
\unfounded_1=\{X\mid &X=\Delta'\cup X'\mbox{, where }\Delta'\subseteq(I_1\setminus \stateI{S_0}), X'\in\stateunfounded{S_0}\mbox{, and }\\&
                     r\mbox{ is not an external support of }X\mbox{with respect to }I_1\}.
\end{array}
\]
$S_1$ is a successor of state $S_0$,
therefore $S_1$ is also a state by Corollary~\ref{cor:successorstate}.
As $\progC_1\subseteq\progC$, $I_1\subseteq I$, $I \cap \negpart{I_1}=\emptyset$,
and 
$\abs{\reductFLP{\progC}{\inter}\setminus\progC_1}=i$,
by the induction hypothesis, $S_1,\dots,S_n$ is a computation, where
$S_n$ is a stable state, $\stateprog{S_n}=\reductFLP{\progC}{\inter}$, and $\stateI{S_n}=I$.
Since $S_1$ is a successor of state $S_0$,
also $S_0,S_1,\dots,S_n$
is a computation.
\ep

For establishing Theorem~\ref{th:unfoundedfreecomp} we make use of the following notion
which reflects positive dependency on the rule level.
\bd{def:ruleDepGraph}
The \memph{positive rule dependency graph} of \progC is given by
\[\rdepG{\progC}=\tuple{\progC,\{\tuple{r_1,r_2}\mid r_1,r_2 \in \progC, \posOcc{\body{r_1}}\cap\posOcc{\head{r_2}}\neq\emptyset\}}.\]
\ed

We can relate the two notions of dependency graph as follows.
\begin{lemma}\label{lemma:acycifacyc}
Let \progC be a \cprogram.
\rdepG{\progC} is acyclic iff \depG{\progC} is acyclic.
\end{lemma}
\bp
Let $\depGrel$ denote the edge relation of \depG{\progC} and
$\rdepGrel$ that of \rdepG{\progC}.

$(\Rightarrow)$
Assume \depG{\progC} is not acyclic.
There must be some path $a_1,\dots,a_n$ of atoms $a_i$ such that for $1\leq i < n$, we have
$a_i\in\domain{\progC}$, $a_i\depGrel a_{i+1}$, and $a_1=a_n$.
Hence, by the definition of $\depG{\progC}$, there must be a sequence
$r_1,\dots,r_{n-1}$ such that for each $1\leq i \leq n-1$, $r_i\in\progC$,
$a_i\in\posOcc{\head{r_i}}$, and
$a_{i+1}\in\posOcc{\body{r_i}}$.
Therefore, for each $1\leq i < n-1$, we have $r_{i+1}\rdepGrel r_i$.
Note that $a_{1}\in\posOcc{\head{r_{1}}}$ and $a_{1}\in\posOcc{\body{r_{n-1}}}$.
Consequently, we have $r_{n-1}\rdepGrel r_1$ and thus $r_{1},r_{n-1},\dots,r_1$
forms a cycle in $\rdepG{\progC}$.
It follows that \rdepG{\progC} is not acyclic.

$(\Leftarrow)$
Assume now that \rdepG{\progC} is not acyclic.
There must be some path $r_1,\dots,r_n$ of \crules $r_i$ such that for $1\leq i < n$ we have
$r_i\in\progC$, $r_1=r_n$, and $r_i\rdepGrel r_{i+1}$.
Hence, by the definition of $\rdepG{\progC}$, there must be a sequence
$a_1,\dots,a_{n-1}$ such that for each $1\leq i \leq n-1$,
$a_i\in\posOcc{\head{r_{i+1}}}$, and
$a_i\in\posOcc{\body{r_i}}$.
Therefore, for each $1\leq i < n-1$ we have $a_{i+1}\depGrel a_i$.
Note that 
$a_{n-1}\in\posOcc{\head{r_{1}}}$ and
$a_1\in\posOcc{\body{r_1}}$.
Consequently, we have $a_{n-1}\depGrel a_1$ and thus $a_{1},a_{n-1},\dots,a_1$
forms a cycle in $\depG{\progC}$.
We have that \rdepG{\progC} is not acyclic.
\ep

\begin{lemma}\label{lemma:topOrder}
Let $\progC$ be an absolutely tight \cprogram.
There is a strict total order $\topOrder$ on $\progC$ that 
extends the reachability relation of $\rdepG{\progC}$.
\end{lemma}
\bp
By Definition~\ref{def:abstight}, 
\depG{\progC} is acyclic.
Hence, by Lemma~\ref{lemma:acycifacyc}, \rdepG{\progC} is also acyclic.
The conjecture holds, since every directed acyclic tree has a topological ordering.
\ep
We now have the means to show Theorem~\ref{th:unfoundedfreecomp},
guaranteeing the existence of stable computations.
\begin{theoremnr}{\ref{th:unfoundedfreecomp}}
Let $\computation\!=\!S_0,\dots,S_n$ be a computation
such that $S_0$ and $S_n$ are stable and
$\progC_\Delta=\stateprog{S_n}\setminus\stateprog{S_0}$ is a normal, convex, and absolutely tight \cprogram.
Then, there is a stable computation $\computation'\!=\!S'_0,\dots,S'_n$
such that
$S_0=S'_0$ and $S_n=S'_n$.
\end{theoremnr}
\bp
Let \topOrder be the strict total order 
extending the reachability relation of $\rdepG{\progC_\Delta}$ that is guaranteed to exist by Lemma~\ref{lemma:topOrder}.
Let $r(\cdot):\{1,\dots,n\}\mapsto\progC_\Delta$ denote the one-to-one mapping from the integer interval $\{1,\dots,n\}$ to the \crules
from $\progC_\Delta$ such that for all $i,j$ in the range of $r(\cdot)$, we have that $i<j$ implies $r(j) \topOrder r(i)$.
Consider the sequence $\computation'\!=\!S'_0,\dots,S'_n$, where
$S'_0=S_0$, and for all $0\leq i<n$,
\[P'_{i+1}=P'_{i}\cup\{r(i+1)\},\]
\[\stateI{S'_{i+1}}=\stateI{S'_{i}}\cup(\stateI{S_{n}}\cap\domain{r(i+1)}),\]
\[\statenegI{S'_{i+1}}=\statenegI{S'_{i}}\cup(\statenegI{S_{n}}\cap\domain{r(i+1)}),\quad\mbox{and}\] 
\[\stateunfounded{S'_{i+1}}=\{\emptyset\}.\]
\noindent
Notice that $S'_n=S_n$ and 
$\proj{\stateI{S'_{i+1}}}{\domain{\stateprog{S'_i}}}=\proj{\stateI{S'_{i}}}{\domain{\stateprog{S'_i}}}$, for all $0\leq i<n$.
We show that $\computation'$ is a computation by induction on the length of a subsequence of $\computation'$.

As base case consider the sequence $\computation''\!=\!S'_0$.
As $S'_0=S_0$ and $S_0$ is a state, $\computation''$
is a computation.
For the induction hypothesis, assume that for some arbitrary but fixed $i$ with $0\leq i < n$,
the sequence $S'_0,\dots,S'_i$ is a computation.

In the induction step it remains to be shown that $S'_{i+1}$ is a successor of $S'_i$.
Clearly, $S'_{i+1}$ is a state structure, and by definition of $\computation'$, since $\computation$ is a computation and 
\[\proj{\stateI{S'_{i+1}}}{\domain{\stateprog{S_{i+1}}}}=\proj{\stateI{S_{n}}}{\domain{\stateprog{S_{i+1}}}},\]
Conditions~(i), (ii), (iii), and (v) of Definition~\ref{def:successor} for being a successor of $S'_i$ are fulfilled by $S'_{i+1}$.
Let $\Delta$ denote $\stateI{S'_{i+1}}\setminus \stateI{S'_{i}}$.

Next we show that Condition~(iv) holds, \iec
$\stateI{S'_{i}}\models \body{r(i+1)}$.
Note that since Condition~(v) holds, we have $\stateI{S'_{i+1}}\models \body{r(i+1)}$ and hence
(iv) holds in the case $\Delta=\emptyset$.
Towards a contradiction assume $\Delta\neq\emptyset$ and $\stateI{S'_{i}}\not\models \body{r(i+1)}$.
We define $\Delta_{{B^{+}}}=\Delta\cap\posOcc{\body{r(i+1)}}$.

First, consider the case that $\Delta_{{B^{+}}}=\emptyset$.
As $\stateI{S'_{i}}\not\models \body{r(i+1)}$, there must be some \cliteral $L\in\body{r(i+1)}$ such that $\stateI{S'_{i}}\not\models L$.
We know that $\stateI{S'_{i+1}}\models L$.
Consequently,
$\proj{\stateI{S'_{i}}}{\domain{L}}\subset \proj{\stateI{S'_{i+1}}}{\domain{L}}$ and therefore $\proj{\Delta}{\domain{L}}\neq\emptyset$.
Moreover, from $\stateI{S'_{i+1}}\models L$ we have 
\[\proj{\stateI{S'_{i+1}}}{\domain{L}}\subseteq\posOcc{\body{r(i+1)}}.\]
It follows that 
$\proj{\Delta}{\domain{L}}\cap\posOcc{\body{r(i+1)}}\neq\emptyset$, indicating a contradiction to $\Delta_{{B^{+}}}=\emptyset$.
It holds that $\Delta_{{B^{+}}}\neq\emptyset$. 
Note that $X\subseteq \stateI{S_{n}}$.
From that, since $S_n$ is a state,
there must be some \crule $r_{\Delta_{{B^{+}}}}\in\stateprog{S_{n}}$ such that $r_{\Delta_{{B^{+}}}}$ is an external support for $\Delta_{{B^{+}}}$ with respect to $\stateI{S_{n}}$.
It cannot be the case that $r\in\stateprog{S_{0}}$, since $\Delta_{{B^{+}}}\cap \stateI{S'_{i}}=\emptyset$,
therefore, $r_{\Delta_{{B^{+}}}}\in\progC_\Delta$.
As $r_{\Delta_{{B^{+}}}}$ is an external support for $\Delta_{{B^{+}}}$ with respect to $\stateI{S_{n}}$,
for $\{A\}=\head{r_{\Delta_{{B^{+}}}}}$, we have $\stateI{S_{n}}\models A$ and $\proj{\Delta_{{B^{+}}}}{\domain{A}}\neq\emptyset$.

Consider the case that $r_{\Delta_{{B^{+}}}}=r(i+1)$.
From that we get $\posOcc{\head{r(i+1)}}\cap\Delta_{{B^{+}}}\neq\emptyset$.
This, in turn, implies $\posOcc{\head{r(i+1)}}\cap\posOcc{\body{r(i+1)}}\neq\emptyset$ which is a contradiction to 
$\rdepG{\progC_\Delta}$ being acyclic. The latter is guaranteed by absolute tightness of $\progC_\Delta$ and Lemma~\ref{lemma:acycifacyc}.

Consider the case that $r(i+1)\topOrder r_{\Delta_{{B^{+}}}}$.
Then, by definition of $\computation'$ we have that $r_{\Delta_{{B^{+}}}}\in\stateprog{S'_{i}}$.
Hence, from $\proj{\Delta_{{B^{+}}}}{\domain{A}}\neq\emptyset$ follows 
\[\proj{\Delta_{{B^{+}}}}{\domain{\stateprog{S'_{i}}}}\neq\emptyset\qquad\mbox{and thus}\qquad\proj{\stateI{S'_{i+1}}\setminus \stateI{S'_{i}}}{\domain{\stateprog{S'_{i}}}}\neq\emptyset.\]
\vspace{-2pt}
$\mbox{The latter is a contradiction to }\proj{\stateI{S'_{i+1}}}{\domain{\stateprog{S'_{i}}}}=\proj{\stateI{S'_{i}}}{\domain{\stateprog{S'_{i}}}}$.

Consider the remaining case that $r_{\Delta_{{B^{+}}}}\topOrder r(i+1)$.
As $\proj{\Delta_{{B^{+}}}}{\domain{A}}\neq\emptyset$, $\Delta_{{B^{+}}}\subseteq \stateI{S_{n}}$, and $\proj{\stateI{S_{n}}}{\domain{A}}\in\satisfiers{A}$, it holds that
$
\posOcc{\head{r_{\Delta_{{B^{+}}}}}}\cap\Delta_{{B^{+}}}\neq\emptyset$.
Therefore, we have $\posOcc{\head{r_{\Delta_{{B^{+}}}}}}\cap\posOcc{\body{r(i+1)}}\neq\emptyset$.
This implies $r(i+1)\topOrder r_{\Delta_{{B^{+}}}}$, being a contradiction to $\topOrder$ being a strict order as we also have $r_{\Delta_{{B^{+}}}}\topOrder r(i+1)$.
Thus, Condition~(iv) of Definition~\ref{def:successor} for being a successor of $S'_i$ holds for $S'_{i+1}$.

Towards a contradiction assume Condition~(vi) does not hold.
Hence, it must hold that 
there is some $\Delta'\subseteq\Delta$ such that $\Delta'\neq\emptyset$ and
$r(i+1)$ is not an external support for $\Delta'$ with respect to $\stateI{S'_{i+1}}$.
We have $\stateI{S'_{i+1}}\models \body{r(i+1)}$ and since we already know that $\stateI{S'_{i}}\models \body{r(i+1)}$, also $\stateI{S'_{i+1}}\setminus \Delta'\models \body{r(i+1)}$ holds by convexity of $\progC_\Delta$.
Moreover, as $\stateI{S'_{i+1}}\models r(i+1)$, it must hold that $\stateI{S'_{i+1}}\models A$ for $\head{r(i+1)}=\{A\}$.
Consequently, for $r(i+1)$ not to be an external support for $\Delta'$ with respect to $\stateI{S'_{i+1}}$, we have $\proj{\Delta'}{\domain{A}}=\emptyset$.
As then $\proj{\Delta'}{\domain{\head{r(i+1)}}}=\emptyset$ but
$\proj{\Delta'}{\domain{r(i+1)}}\neq\emptyset$ it must hold that 
$\proj{\Delta'}{\domain{\body{r(i+1)}}}\neq\emptyset$.
Consider $\Delta''=\Delta'\cap\posOcc{\body{r(i+1)}}$ and assume that $\Delta''\neq\emptyset$.
Then, as $\Delta''\subseteq \stateI{S_{n}}$, there must be some \crule $r_{\Delta''}$ that is an external support for $\Delta''$ with respect to $\stateI{S_{n}}$.
Hence, $\posOcc{\head{r_{\Delta''}}}\cap\Delta''\neq\emptyset$ and therefore 
$\posOcc{\head{r_{\Delta''}}}\cap\posOcc{\body{r(i+1)}}\neq\emptyset$.
It follows that $r(i+1)\topOrder r_{\Delta''}$. From that we get $r_{\Delta''}\in\stateprog{S'_{i}}$.
This is a contradiction 
as we know that $\posOcc{\head{r_{\Delta''}}}\cap\Delta''\neq\emptyset$,
$\posOcc{\head{r_{\Delta''}}}\cap\Delta''\subseteq \stateI{S'_{i}}$,
and $\Delta''\subseteq \stateI{S'_{i+1}} \setminus \stateI{S'_{i}}$.
Consequently, $\Delta'\cap\posOcc{\body{r(i+1)}}=\emptyset$ must hold.
From $\proj{\Delta'}{\domain{\body{r(i+1)}}}\neq\emptyset$
we get that there is some $L\in\body{r(i+1)}$ with $\proj{\Delta'}{\domain{L}}\neq\emptyset$.
As $\stateI{S'_{i+1}}\models L$, we have that $\proj{\stateI{S'_{i+1}}}{\domain{L}}\in C$ in the case $L$ is a \catom $L=\tuple{\domain{L},C}$, and $\proj{\stateI{S'_{i+1}}}{\domain{L}}\in 2^\domain{L}\setminus C$ in the case $L$ is a default negated \catom $L=\naf\ \tuple{\domain{L},C}$. In both cases,
as $\Delta'\subseteq \stateI{S'_{i+1}}$ and $\proj{\Delta'}{\domain{L}}\neq\emptyset$,
we get a contradiction to $\Delta'\cap\posOcc{\body{r(i+1)}}=\emptyset$.
\ep \fi

\bibliographystyle{acmtrans}

\end{document}